\documentclass[runningheads]{llncs}

\usepackage{eccv}

\usepackage{eccvabbrv}

\usepackage{graphicx}
\usepackage{booktabs}

\usepackage[accsupp]{axessibility}  % Improves PDF readability for those with disabilities.

\usepackage{booktabs}
\usepackage{multirow}
\usepackage[table]{xcolor}
\usepackage{array}
\usepackage{array}
\usepackage{subcaption}
\usepackage{xspace}
\usepackage{kotex}

\usepackage{graphicx}
\usepackage{stfloats}
\usepackage{caption}

\usepackage{amsmath}
\usepackage{amssymb}
\usepackage{wrapfig}
\usepackage{float}
\usepackage{algorithm}

\usepackage{verbatim}

\usepackage{algorithm}
\usepackage{algpseudocode}
\usepackage{chngcntr}
\usepackage{float}

\definecolor{pearDark}{RGB}{190,215,84}
\definecolor{mycolor_green}{RGB}{200,230,200}
\definecolor{mycolor_yellow}{RGB}{255,240,180}

\usepackage{hyperref}

\usepackage{orcidlink}
\newcommand{\conceptmix}{\textsc{ConceptMix}\xspace}
\newcommand{\fullname}{Correlation-Weighted Multi-Reward Optimization\xspace}
\newcommand{\ours}{CMO\xspace}

\renewcommand{\footnoterule}{%
  \kern -3pt
  \hrule width 0.4\textwidth height 0.4pt
  \kern 2.6pt
}

\begin{document}

\title{Correlation-Weighted Multi-Reward Optimization for Compositional Generation} 

\titlerunning{CMO}
% Second Author\inst{2,3}\orcidlink{1111-2222-3333-4444} \and
% Third Author\inst{3}\orcidlink{2222--3333-4444-5555}}

\author{Jungmyung Wi\inst{1}\orcidlink{0009-0005-0913-6566} \and
Hyunsoo Kim\inst{1,2}\orcidlink{0009-0006-7456-2876}\and
Donghyun Kim$^{\dagger}$\inst{1}\orcidlink{0000-0002-7132-4454}}

\authorrunning{J.~Wi et al.}

\institute{Korea University \and
    Korea Institute of Science and Technology \\
    \email{\{wjm333, climba, d\_kim\}@korea.ac.kr}
    }

\maketitle

\begingroup
\renewcommand{\thefootnote}{\textdagger}
\NoHyper
\footnotetext{Corresponding author.}
\endNoHyper
\endgroup

\begin{abstract}

Text-to-image models produce images that align well with natural language prompts, but compositional generation has long been a central challenge. Models often struggle to satisfy multiple concepts within a single prompt, frequently omitting some concepts and resulting in partial success. Such failures highlight the difficulty of jointly optimizing multiple concepts during reward optimization, where competing concepts can interfere with one another. 
To address this limitation, we propose Correlation-Weighted Multi-Reward Optimization (\ours), a framework that leverages the correlation structure among concept rewards to adaptively weight each attribute concept in optimization. By accounting for interactions among concepts, \ours balances competing reward signals and emphasizes concepts that are partially satisfied yet inconsistently generated across samples, improving compositional generation.
Specifically, we decompose multi-concept prompts into pre-defined concept groups (\eg, objects, attributes, and relations) and obtain reward signals from dedicated reward models for each concept. We then adaptively reweight these rewards, assigning higher weights to conflicting or hard-to-satisfy concepts using correlation-based difficulty estimation. By focusing optimization on the most challenging concepts within each group, \ours encourages the model to consistently satisfy all requested attributes simultaneously. We apply our approach to train state-of-the-art diffusion models, SD3.5 and FLUX.1-dev, and demonstrate consistent improvements on challenging multi-concept benchmarks, including ConceptMix, GenEval 2, and T2I-CompBench. The code is available at \url{https://github.com/TheDarkKnight-21th/CMO}.

\end{abstract}    
\section{Introduction}
\label{sec:intro}

\begin{figure*}[t]
    \centering
    \includegraphics[width=0.9\linewidth]{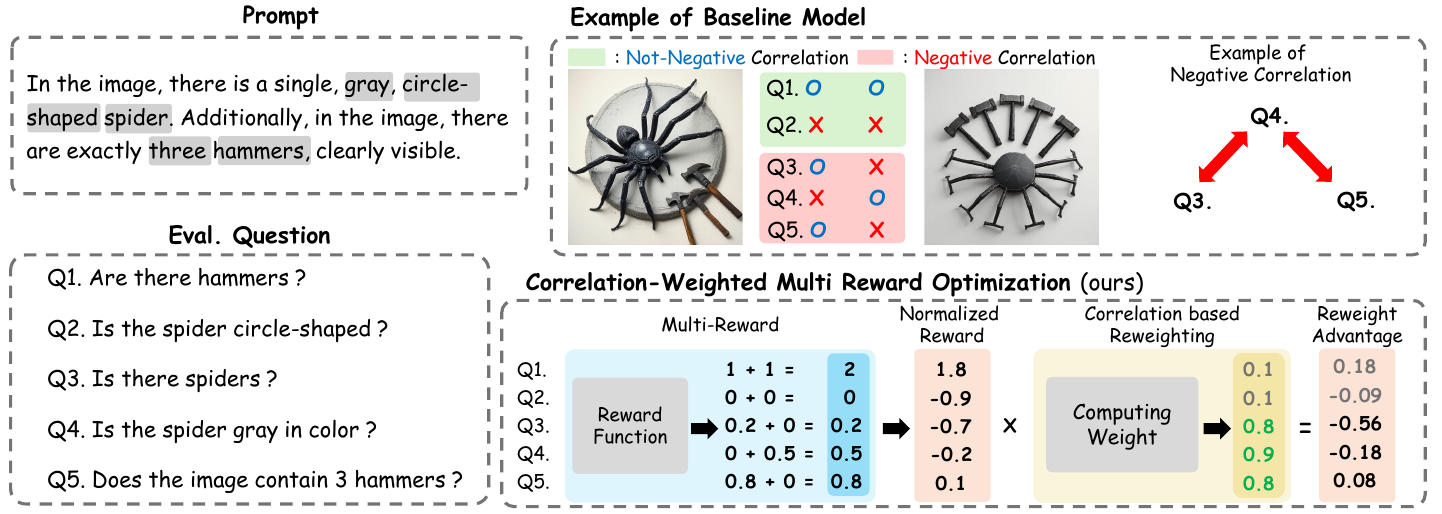} 
    \caption{\textbf{Motivation of Correlation-based Reweighting.} Multi-concept compositional generation often yields different partial successes distributed across generated images. Thus, naively aggregating rewards from each concept tends to focus on concepts that are consistently satisfied across images, while failing to highlight more difficult concepts. Correlation computes concept interactions among generated instances to identify difficult concepts, which are then assigned higher weights during normalized advantage reweighting (\ie, Q3, Q4, and Q5).
    }
    \label{fig:teaser}
    % \vspace{-2em}
\end{figure*}
Text-to-image diffusion models~\cite{rombach2022high,podell2023sdxl,chen2023pixart,sauer2024adversarial,esser2024scaling,wu2025qwen,wu2025janus,xie2024show} have become the standard paradigm for text-conditioned image generation, producing images that align well with natural language prompts. Early breakthroughs in this area were primarily driven by UNet-based diffusion models~\cite{rombach2022high,podell2023sdxl}. Recently, the field has rapidly evolved with the introduction of scalable flow-matching based models~\cite{chen2023pixart,sauer2024adversarial,esser2024scaling,wu2025qwen} and autoregressive generative models~\cite{wu2025janus,xie2024show}. As model architectures and training corpora continue to scale, modern generative models are increasingly capable of generating diverse visual concepts described in text with high fidelity.

Despite these advances in generation quality, compositional generation has long been a central challenge. Real-world prompts often require models to align multiple concepts simultaneously while correctly binding each attribute to its intended object and relation. Early generative models struggled with token entanglement, frequently failing when multiple concepts are expressed within a single prompt~\cite{ghosh2023geneval,huang2023t2i}.
Prior works attempted to mitigate this issue through inference-time attention control~\cite{liu2022compositional,feng2022training,chefer2023attend,hu2024token} or attention map manipulation using additional conditioning signals~\cite{dat2025vsc,zhang2024realcompo,dahary2024yourself}. More recent architectures, such as DiT and flow-matching models~\cite{chen2023pixart,peebles2023scalable,esser2024scaling,flux2024,wu2025qwen}, improve token separation and partially alleviate this problem.

To further improve compositional generation, recent diffusion reinforcement learning (RL) approaches~\cite{liu2025flow,yang2025hermesflow,li2025mixgrpo,he2508tempflow,Pref-GRPO&UniGenBench} have successfully optimized a simple concept generation using evaluation metrics (\eg, GenEval~\cite{ghosh2023geneval}) as rewards. However, applying this RL paradigm to multi-concept prompts reveals a critical bottleneck. Models frequently fail to generate all concepts together, resulting in partial success (\eg, correct objects but wrong numeracy)~\cite{wu2024conceptmix,kamath2025geneval,li2024genai,wei2025tiif}. Existing approaches~\cite {liu2025flow,yang2025hermesflow,li2025mixgrpo,he2508tempflow,Pref-GRPO&UniGenBench} attempt to solve this by naively aggregating individual rewards for each concept. However, as illustrated in Figure~\ref{fig:teaser}, such naive aggregation tends to overlook difficult or negatively correlated concepts (\ie, Q3, Q4, and Q5). For example, it may overemphasize easier concepts (Q1), while optimizing for the gray color (Q4) often leads to the omission of the spider (Q3) or the hammers (Q5). Consequently, the model prioritizes easier concepts and fails to generate all requested attributes simultaneously.

To address this, we propose \textbf{C}orrelation-Weighted \textbf{M}ulti-Reward \textbf{O}ptimization (\textbf{\ours}), which adaptively reweights concept-wise rewards by leveraging the correlation structure among reward signals across generated samples.
% Prior work typically overlooks the inherent differences in generative difficulty among concepts, instead relying on a naive aggregation of their rewards. This causes optimization to be dominated by easily satisfied concepts, while more challenging concepts remain neglected.
Since difficulty varies depending on the specific combination of concepts in each prompt, manual weight adjustment becomes impractical. To address this, \ours~ automatically estimates concept difficulty within each sampled group. By assigning higher weights to these hard-to-align concepts (\eg, `three hammers' in Figure~\ref{fig:teaser}), \ours encourages the model to generate all requested concepts together. To enable this optimization, we first design a \textit{multi-concept reward} that decomposes multi-concept prompts into fine-grained objectives covering objects, attributes, numeracy, and spatial relations. Each concept is evaluated with dedicated reward functions, producing a set of concept-wise rewards that reflect different aspects of compositional correctness. These rewards are then used by \ours~to estimate concept interactions and dynamically reweight difficult concepts.
For optimization, we build on recent reward optimization recipes that employ group-wise comparisons for stable learning. In particular, we adopt MixGRPO~\cite{li2025mixgrpo} for efficient mixed ODE--SDE sampling and GDPO~\cite{liu2026gdpo} for reward-decoupled normalization across multiple rewards. 

We apply \ours to SD3.5~\cite{esser2024scaling} and FLUX.1-dev~\cite{flux2024}. Extensive experiments demonstrate consistent improvements on multi-concept benchmarks such as ConceptMix~\cite{wu2024conceptmix} and Geneval~2~\cite{kamath2025geneval}, as well as on T2I-CompBench~\cite{huang2023t2i}, which evaluates fine-grained attribute binding. Our contributions can be summarized as follows:
\begin{itemize}
\item We identify a key limitation of existing RL approaches for multi-concept compositional generation: naive aggregation of concept-wise rewards overemphasizes easy concepts while overlooking difficult or negatively correlated ones.
\item We propose Correlation-Weighted Multi-Reward Optimization (\ours), which estimates concept difficulty from correlations among reward signals across generated samples and adaptively reweights concept-wise rewards to emphasize hard-to-satisfy concepts.
\item We apply \ours~ to train state-of-the-art diffusion models (SD3.5 and FLUX.1-dev), demonstrating consistent improvements on challenging multi-concept benchmarks (\eg, ConceptMix, GenEval~2, and T2I-CompBench).
\end{itemize}

\section{Related Work}
\label{sec:related_work}

\subsection{Compositional Generation in T2I Models}

Compositional generation has long been recognized as a central challenge in text-to-image (T2I) diffusion models, where early models struggle to render multiple concepts such as objects, attributes, and spatial relations expressed in a single prompt~\cite{ghosh2023geneval,huang2023t2i}. To address these issues, prior works have focused on improving attribute-level disentanglement through attention manipulation, token interaction control, or additional conditioning mechanisms~\cite{liu2022compositional,feng2022training,chefer2023attend,hu2024token,dat2025vsc,zhang2024realcompo}. While architectural advancements such as DiT and flow-matching models~\cite{peebles2023scalable,esser2024scaling,flux2024} have largely mitigated initial entanglement problems, generating multiple concept together remains challenging~\cite{wu2024conceptmix,kamath2025geneval,wei2025tiif}. Recent efforts have explored diffusion RL to directly optimize reward signals that reflect multi-constraint satisfaction~\cite{zhang2024itercomp,Pref-GRPO&UniGenBench}. However, prior works fail to generate images that satisfy all attributes in a multi-concept prompt simultaneously because they do not consider the relationships between individual attributes within multiple different concepts. To prevent such failures, we propose \ours, a multi-reward optimization framework specifically designed to satisfy multiple concepts simultaneously. By automatically focusing on the hardest concepts, our method ensures that every requested concept is generated at the same time.

\subsection{Reinforcement Learning for Diffusion Model}
Reinforcement learning is widely used to align diffusion/flow-based T2I models~\cite{esser2024scaling,flux2024,wu2025qwen} by optimizing rewards along the denoising process under distributional constraints. Prior work includes PPO-style policy gradient methods with KL regularization (e.g., DDPO~\cite{ddpo} and DPOK~\cite{dpok}), diffusion-adapted preference learning from win--lose pairs without explicit reward-model training (e.g., Diffusion-DPO~\cite{wallace2024diffusion}), sampler-based optimization with differentiable rewards (e.g., DRaFT~\cite{clark2023directly} and AlignProp~\cite{prabhudesai2023aligning}), and scalable black-box reward fine-tuning over large prompt sets (e.g., PRDP~\cite{deng2024prdp}). Recent work extends RL alignment to flow-matching and rectified-flow backbones using group-relative optimization and stochastic exploration to improve compositional correctness and preference alignment, as exemplified by Flow-GRPO~\cite{liu2025flow}, DanceGRPO~\cite{xue2025dancegrpo}, MixGRPO~\cite{li2025mixgrpo} and TempGRPO~\cite{he2508tempflow}. However, these methods simply average multiple reward signals without considering the relationships between them, which can cause optimization to be dominated by easily satisfied concepts while neglecting more difficult or conflicting ones. To address this, we propose \ours, which dynamically reweights multiple rewards to encourage the model to satisfy them simultaneously.

\begin{figure*}[t!]
    \centering
    \includegraphics[width=\textwidth]{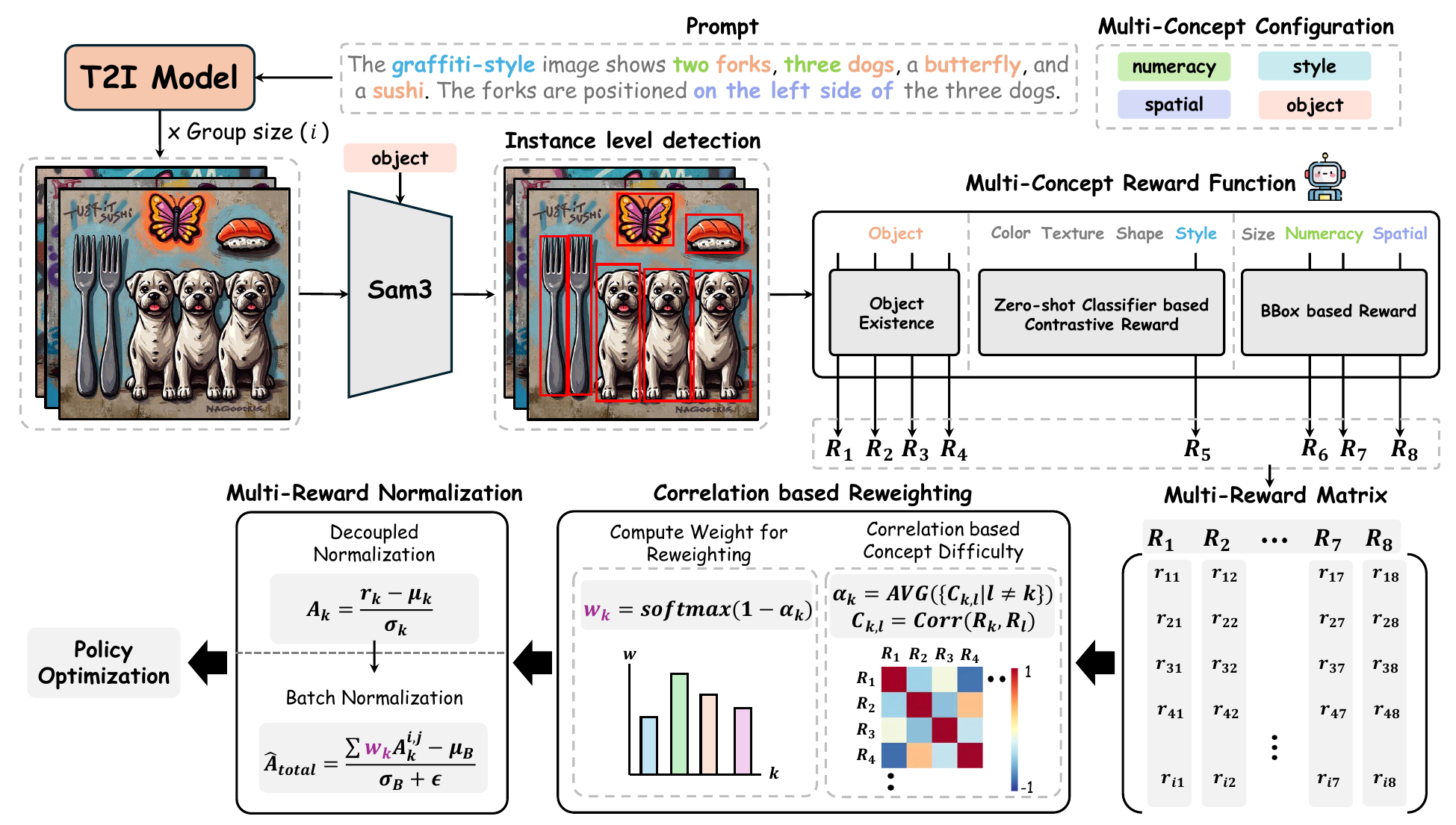}
    \caption{
        \textbf{Overview of Correlation-Weighted Multi-Reward Optimization (\ours).} Given a multi-concept prompt, the text-to-image (T2I) model generates a group of images, which are evaluated using a set of concept-specific reward functions covering objects, attributes, and spatial relations. 
        The resulting concept rewards form a Multi-Reward Matrix across generated instances.
         \ours then computes correlations among reward signals to estimate concept difficulty based on their interactions, assigning higher weights to concepts that are harder to consistently satisfy. The reweighted rewards are normalized and aggregated to guide policy optimization, encouraging the model to jointly satisfy all requested concepts and improving compositional generation.
    }
    % \vspace{-1em}
    \label{fig:overall_pipeline}
\end{figure*}

\section{Method}

\label{sec:method}

In this work, we propose \fullname~ (\ours), a new reward optimization framework tailored for multi-concept compositional generation. The overall pipeline is illustrated in Figure~\ref{fig:overall_pipeline}. First, we present the preliminaries in Sec.~\ref{sec:prelim}. In Sec.~\ref{complex_reward}, we design a multi-concept reward, which decomposes the multi-concept prompt into fine-grained objectives covering objects, attributes, numeracy, and spatial relations, and evaluates each concept with dedicated reward functions. And then in Sec.~\ref{reweight}, we introduce a correlation-based reweighting mechanism that dynamically assigns higher weights to hard-to-align, conflicting concepts to ensure balanced optimization and encourage the model to satisfy all requested concepts simultaneously.

\subsection{Preliminaries}
\label{sec:prelim}

\noindent\textbf{Mixed ODE-SDE Sampling.} To mitigate the computational overhead of full-step SDE sampling, MixGRPO~\cite{li2025mixgrpo} restricts stochastic exploration to a sliding window $S \subseteq [0, T]$. It employs SDE sampling within $S$ for RL exploration, and deterministic ODE sampling elsewhere:
\begin{equation}
    dx_t = \begin{cases}
    [v_t - \frac{1}{2}g^2(t)s_t]dt + g(t)dw, & t \in S \\
    v_t dt, & t \notin S
    \end{cases}
\end{equation}
where $v_t$ and $s_t$ are the velocity field and score function. This concentrates gradients within $S$, significantly shortening the optimization timestep.

\noindent\textbf{Multi-Reward Normalization.} GRPO~\cite{shao2024deepseekmath} often suffers from reward collapse in multi-reward settings, mapping distinct reward combinations to identical advantages due to pre-normalization aggregation. GDPO~\cite{liu2026gdpo} resolves this by decoupling the normalization process. For $K$ rewards, the group-relative advantage $A_k^{(i,j)}$ for the $k$-th reward is computed independently, then weighted, summed, and batch-normalized to obtain the final advantage $\hat{A}_{total}^{(i,j)}$:
\begin{equation}
    A_k^{(i,j)} = \frac{r_k^{(i,j)} - \mu_k^{(i)}}{\sigma_k^{(i)}}, \quad \hat{A}_{total}^{(i,j)} = \frac{\sum A_k^{(i,j)} - \mu_B}{\sigma_B + \epsilon}
\label{advatage}
\end{equation}
where $\mu_k^{(i)}, \sigma_k^{(i)}$ are the group-level mean and standard deviation for the $k$-th reward, and $\mu_B, \sigma_B$ are the batch-level statistics of the weighted sum.

\subsection{Multi-Concept Reward Decomposition}
\label{complex_reward}
To evaluate compositional generation, we decompose a multi-concept prompt into a set of fine-grained concept objectives. Instead of computing a single scalar reward, our framework evaluates multiple concept-level rewards corresponding to objects, attributes, numeracy, and spatial relations. This decomposition allows each concept in the prompt to be evaluated independently, providing more informative supervision for reward optimization.
Formally, given a generated image $x$ and a multi-concept configuration $\mathcal{C}$ (\eg,`style':`graffti', `object':`dog', `numeracy':`two' in Figure~\ref{fig:overall_pipeline}) extracted from the prompt, we compute a set of concept-wise rewards. Each reward corresponds to a specific aspect of compositional correctness. Detailed formulations and implementation details for each reward component are provided in the Appendix~\ref{sec:multi_reward}.

\noindent\textbf{Object Existence Reward.}
For each object specification in $\mathcal{C}$, we extract the specific text corresponding to the `object' key in $\mathcal{C}$ and feed it as a prompt into a segmentation model (\eg, SAM 3~\cite{carion2025sam}) to perform text-guided instance object detection. 
Let $\mathcal{O}_i$ denote the set of detected instances for object $i$. For each valid instance, we extract its bounding box $b$ and segmentation mask $m$. 
Based on these detections, we assign a binary existence reward:
\begin{equation}
r^{\text{exist}}_i = 
\begin{cases}
1, & \text{if } |\mathcal{O}_i| > 0, \\
0, & \text{otherwise}.
\end{cases}
\end{equation}
Crucially, the extracted bounding boxes and masks serve as the precise foundation for subsequent localized evaluations, effectively isolating entities from background noise.

\noindent\textbf{Zero-shot Classifier based Contrastive Reward for Attribute Reward.}
\label{attribute_level_reward}
For visual attributes such as color, texture, shape, and style, we measure semantic alignment using a zero-shot classifier (\ie, OpenCLIP-H~\cite{cherti2023reproducible}). To isolate the target object from background noise, we extract the image feature $f_{\text{img}}(x \odot b_i)$ from the cropped region defined by the bounding box $b_i$, where $x$ is the generated image and $b_i$ is the bounding box of the target object. For each attribute category $k$ (e.g., color), we maintain a predefined candidate set $\mathcal{V}_k$ (e.g., {red, green, blue}). Let $f_v$ denote the normalized text feature corresponding to candidate attribute $v \in \mathcal{V}_k$. We compute similarities between the image feature and all candidate attributes, and use a softmax classifier to estimate the probability of the target attribute $v^*$ specified in the prompt. The attribute reward is defined as:
\begin{equation}
\label{eq:attr_reward}
r_i^{\text{attr}}
=
\frac{
\exp \left(
\langle f_{\text{img}}, f_{v^*} \rangle
\right)
}
{
\sum_{v \in \mathcal{V}_k}
\exp \left(
\langle f_{\text{img}}, f_v \rangle
\right)
}.
\end{equation}
Additional details on the construction and normalization of the candidate text features are provided in the Appendix~\ref{sec:multi_reward}.

\noindent\textbf{BBox based Reward.}
For numerical reward, instead of a binary penalty such as object existence, we propose a differentiable heuristic based on count deviation. Given the detected count $c_i = |\mathcal{O}_i|$ and the target count $c^*_i$, the reward is formulated as:
\begin{equation}
r^{\text{num}}_i = \frac{1}{(|c_i - c^*_i| + 1)^2}.
\end{equation}

Similarly, size attributes (\eg, {huge}, {tiny}) are evaluated using a rank-based inverse squared decay over the areas of all detected objects in the scene.
For specified spatial relations between objects $i$ and $j$, we formulate geometric heuristics to assign partial scores ($1.0$, $0.5$, or $0.0$) based on alignment constraints.

Notably, for 3D depth relations (\eg, in front of, behind), we leverage the depth estimation from Depth Anything 3~\cite{lin2025depth} to extract a normalized depth map $\mathcal{D}(x)$. We compute the average depth $\bar{d}_i$ within the masked region $m_i$:
\begin{equation}
\bar{d}_i = \frac{1}{|m_i|} \sum_{p \in m_i} \mathcal{D}(x)[p].
\end{equation}
The depth-based relation reward seamlessly incorporates a tolerance margin $\epsilon$ to account for estimation noise. For instance, the reward for $i$ being \textit{in front of} $j$ is formulated as:
\begin{equation}
\label{eq:3d_spatial}
r^{\text{rel}}_{i,j} = 
\begin{cases} 
1.0, & \text{if } \bar{d}_i < \bar{d}_j - \epsilon, \\ 
0.5, & \text{if } \bar{d}_i < \bar{d}_j, \\ 
0.0, & \text{otherwise}.
\end{cases}
\end{equation}
Conversely, the reward for the \textit{behind} relation is evaluated using the exact opposite condition. Detailed formulations for 2D spatial relations (\eg, above, below, to the left of, to the right of) are provided in the Appendix~\ref{sec:multi_reward}.

\subsection{Correlation-based Reweighting for Multi-Reward Optimization}
\label{reweight}

\subsubsection{Motivation.} 
\begin{figure}[t]
\centering
\includegraphics[width=\linewidth]{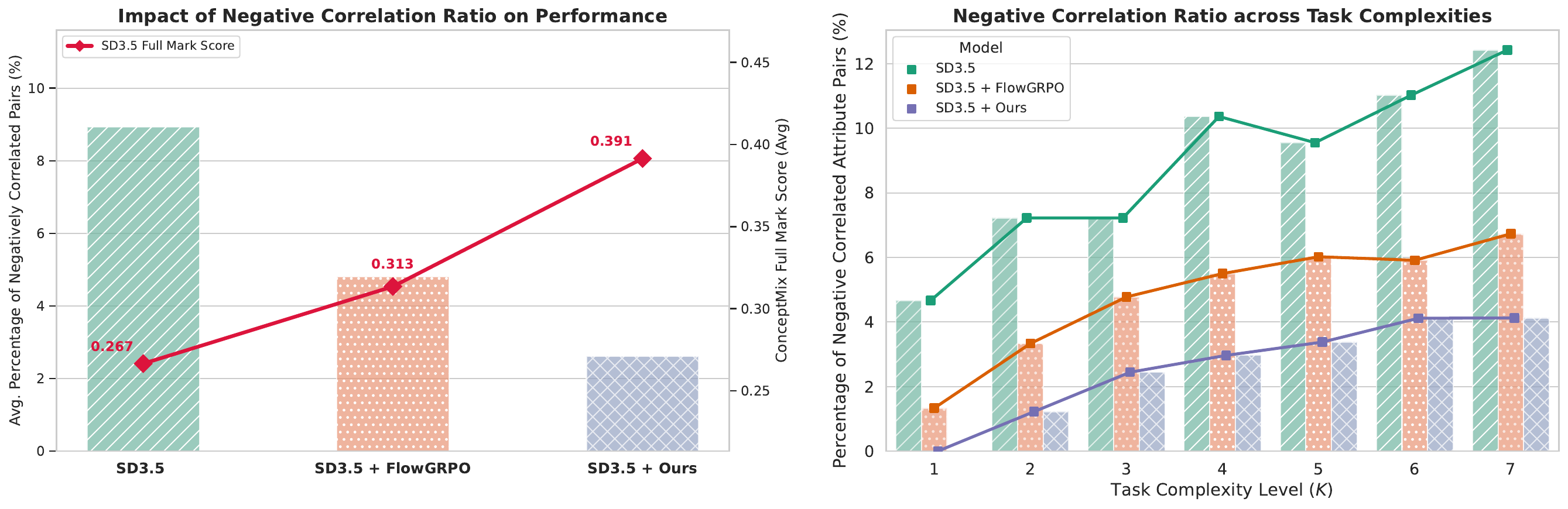} 
\vspace{-2mm}
\caption{\textbf{Quantitative Analysis on Multi-Concept Generation Conflicts.} 
\textbf{Left:} The graph compares the average ratio of negative correlations and the overall Full Mark Score on Conceptmix~\cite{wu2024conceptmix}, showing that reducing negative correlations between concepts directly improves multi-concept generation. \textbf{Right:} The graph tracks the percentage of concept pairs exhibiting negative correlation as the number of concepts (task complexity level $K$) increases. Baselines exhibit increasing negative correlations as complexity grows, whereas our method maintains a low negative correlation ratio with correlation-based reweighting.}
\label{fig:reward_analysis}
\vspace{-1em}
\end{figure}

While the multi-concept reward provides fine-grained supervision for each concept, naively aggregating these rewards can lead to suboptimal optimization. GRPO \cite{lu2024deepseek} aggregates multiple rewards through naive summation, and GDPO~\cite{liu2026gdpo} decouples with reward normalization, and it aggregates advantages. As a result, concepts that are easy to satisfy tend to dominate the optimization process, while difficult or conflicting concepts receive insufficient training signals.
To verify this, we analyze correlations of each concept reward across generated instances.  Using the ConceptMix set, we define a generation group as 10 images synthesized from a single prompt with different random seeds to analyze the relationship between concepts. Figure~\ref{fig:reward_analysis} (left) shows a negative correlation between concepts and the full-mark score on the ConceptMix~\cite{wu2024conceptmix} benchmark across baselines (SD 3.5, SD 3.5 + FlowGRPO, and SD 3.5 + Ours). There is a clear tendency that higher negative correlation leads to lower full-mark scores, indicating a generation conflict where successfully generating one concept limits the ability to generate another. Crucially, Figure~\ref{fig:reward_analysis} (right) shows that baseline models exhibit a sharp increase in the ratio of negatively correlated concept pairs as the number of concepts (prompt complexity, $K$) grows. This observation suggests that naive aggregation exacerbates trade-offs between concepts. However, our correlation-based reweighting shows that prioritizing difficult and negatively correlated concepts can mitigate generation conflicts and significantly improve multi-concept compositional generation even at high levels of complexity.

\noindent\textbf{Correlation-Based Concept Difficulty Estimation.}
For a given prompt $i$, we generate a group of $G$ images. 
In reward optimization for diffusion models, concept-wise rewards are collected to evaluate generation quality, yielding a reward matrix $\mathbf{R}^{(i)} \in \mathbb{R}^{G \times K}$, where $K$ is the number of concepts. 
To stably estimate the difficulty among the concepts, we calculate the Pearson correlation~\cite{sedgwick2012pearson} matrix $\mathbf{C} \in \mathbb{R}^{K \times K}$ across the $K$ concepts over the $G$ samples:
\begin{equation}
C_{k,l} = \text{Corr}(\mathbf{R}_{k}^{(i)}, \mathbf{R}_{l}^{(i)}).
\end{equation}

During fine-tuning, certain concepts may be perfectly satisfied across all samples in a group, resulting in zero variance ($\text{std}(\mathbf{R}_{:,k}^{(i)}) \approx 0$). In such cases, the correlation is mathematically undefined. To ensure training stability, we simply bypass the correlation estimation for these static concepts and assign a default maximum correlation score to prevent unstable gradient. For each concept $k$, we define a score {$\alpha_k$} as the average correlation with other attributes:
\begin{equation}
\alpha_k = 
\begin{cases} 
1.0, & \text{if } \text{std}(\mathbf{R}_{k}^{(i)}) < \epsilon, \\ 
\frac{1}{K-1} \sum_{l \neq k} C_{k,l}, & \text{otherwise}. 
\end{cases}
\end{equation}
Since $C_{k,l} \in [-1, 1]$, negatively correlated (i.e., difficult to generate simultaneously) concepts yield lower scores. To ensure training stability, if the generated group contains incomplete padding values, we safely bypass this correlation estimation.

\noindent\textbf{Correlation-based Reweighting.}
Concepts with high scores are easier to satisfy jointly and therefore receive lower gradient emphasis. 
Conversely, weakly correlated concepts are considered more difficult and receive higher weights. 
We define an correlation-based weighting using a softmax function:
\begin{equation}
w_k = \frac{\exp\big((1 - \alpha_k))}{\sum_{n=1}^{K} \exp\big((1 - \alpha_n))},
\end{equation}

To compute the final advantage for the $j$-th sample in group $i$, we substitute the aggregation in Eq.~\ref{advatage} with our correlation-based weights. The weighted sum is then batch-normalized to obtain the final advantage :
\begin{equation}
\hat{A}_{\text{total}}^{(i,j)} = \frac{\sum w_k A_k^{(i,j)} - \mu_B}{\sigma_B + \epsilon}
\end{equation}

This strategy dynamically prioritizes concepts difficult to generate simultaneously via weighted averaging, encouraging balanced reward optimization.
\section{Experiment}
\label{sec:experiment}

In this section, we evaluate the effectiveness of Correlation-Weighted Multi-Reward Optimization (\ours) for improving compositional generation. We first describe the experimental setup in Sec.~\ref{sec:exp_setup}, including the training configuration and evaluation benchmarks. We then compare \ours with existing diffusion RL baselines across multiple compositional generation benchmarks in Sec.~\ref{sec:main_results}. To further analyze the behavior of \ours, we provide detailed ablations on the correlation-based reweighting mechanism and study its impact on resolving concept conflicts in Sec.~\ref{sec:ablation}. Finally, we present qualitative examples demonstrating improved multi-concept generation under increasing compositional complexity in Sec.~\ref{sec:qual}.

\subsection{Experimental Setup}
\label{sec:exp_setup}
\noindent\textbf{Training Prompt Dataset.}
\label{subsec:dataset_construction}
Our training dataset comprises 5k prompts: 2.5k complex multi-concept and 2.5k single-attribute prompts. For the complex multi-concept set, we synthesize prompts binding up to eight concepts per entity ($K+1=8$) using the ConceptMix~\cite{wu2024conceptmix} pipeline and Qwen3-30B~\cite{yang2025qwen3}. Motivated by recent findings~\cite{liu2025flow,he2508tempflow} that a single-attribute prompt set(\eg, `\{color\} \{object\}', `\{color\} \{object\} and \{color\} \{object\}) enhances overall compositional generation, we include the single-attribute set. To focus the policy on the most challenging attributes, we heavily skew their sampling distribution (Color : Texture : Shape : Size : Spatial : Numeracy) to a 3:3:2:1:10:7 ratio.

\noindent\textbf{Base Models and Training Setup.} 
We fine-tune Stable Diffusion 3.5~\cite{esser2024scaling} and FLUX.1-dev~\cite{flux2024} using our based multi-concept reward optimization with LoRA~\cite{hu2022lora}. Each training iteration samples 8 prompts, generating 16 images per prompt across 4 GPUs, yielding a global batch size of 512. The learning rate is set to $3 \times 10^{-4}$. During training, we use 10 sampling steps for SD3.5 and 6 for FLUX.1-dev. To balance exploration and computational efficiency, we employ a mixed ODE-SDE sampling strategy with Coefficient Preserving Sampling (CPS)~\cite{wang2025coefficients}. Following MixGRPO~\cite{li2025mixgrpo}, stochastic SDE updates are confined strictly to the first 3 timesteps, while the remaining steps rely on deterministic ODE sampling.

\noindent\textbf{Evaluation Protocol.}
\label{sec:evaluation_protocol}
We evaluate our method on three compositional text-to-image benchmarks: ConceptMix~\cite{wu2024conceptmix}, GenEval 2~\cite{kamath2025geneval}, and T2I-CompBench~\cite{huang2023t2i}. To account for generative stochasticity, we synthesize 10 images per prompt using different random seeds across all benchmarks and report the averaged metrics. While following official protocols, we adapt ConceptMix: originally, it relies on GPT-4o~\cite{hurst2024gpt} as its primary evaluator and provides DeepSeek-VL-7B-chat~\cite{lu2024deepseek} as an open-source alternative. To ensure stable reproducibility, we instead employ the more capable Qwen3-VL-8B~\cite{bai2025qwen3}. All baselines are strictly re-evaluated using this exact setup for fair comparison.

\noindent\textbf{Baselines.}
\label{sec:baselines}
We compare our method against a variety of representative compositional T2I baselines. These include Qwen-Image~\cite{wu2025qwen} as a strong open-weight generator, and recent RL-based alignment methods such as IterComp~\cite{zhang2024itercomp}, Flow-GRPO~\cite{liu2025flow} on SD3.5, and Pref-GRPO~\cite{Pref-GRPO&UniGenBench} on FLUX.1-dev. Due to evaluation costs, NanoBanana~\cite{comanici2025gemini} is evaluated exclusively on T2I-CompBench. To demonstrate backbone-agnostic scalability, we report results for both SD3.5 and FLUX.1-dev in T2I-CompBench. For other benchmarks, we focus on the high-capacity FLUX.1-dev model to evaluate our method against large-scale baselines such as Qwen-Image.

\begin{table}[ht!]
    \centering
    \small
    \setlength\arrayrulewidth{1pt}
    \setlength\tabcolsep{6pt}
    \caption{\textbf{Full Mark Score of T2I Models on \conceptmix.} We report full mark scores across difficulty levels $k=1$ to $k=7$, where $k$ denotes the number of required concepts (thus each prompt contains $k\!+\!1$ concepts). As $k$ increases, performance degrades for all models, while our method consistently mitigates degradation at higher complexity. \textbf{Bold} and \underline{underline} indicate the best and second-best results, respectively.}
    \label{tab:main}
    \begin{subtable}[t]{\textwidth}
    \centering
    \resizebox{\linewidth}{!}{%
    \begin{tabular}{lccccccc}
    \toprule
              Models   & $k=1$                & $k=2$                & $k=3$                & $k=4$                & $k=5$                & $k=6$                & $k=7$                \\
    \midrule 
    \rowcolor{gray!15}\multicolumn{8}{c}{\textit{Autoregressive Models}} \\
    \midrule 
    Janus~\cite{wu2025janus}  & 0.5340 & 0.3063 & 0.1710 & 0.0713 & 0.0383 & 0.0173  & 0.0083 \\
    Show-o~\cite{xie2024show} & 0.7132 & 0.4560 & 0.2747 &  0.1507 & 0.1100 & 0.0583 & 0.0357 \\
    HermesFlow~\cite{yang2025hermesflow} & 0.6993 & 0.4560 & 0.2690 & 0.1467 & 0.1167 & 0.0540 & 0.0403 \\
    \midrule 
    \rowcolor{gray!15}\multicolumn{8}{c}{\textit{Non-Autoregressive Models}} \\
    \midrule 
    StableDiffusion v1.4~\cite{rombach2022high}        & {0.4780}  & {0.1943}  & {0.0717}    & {0.0203}   & {0.0040}    & {0.0007}   & {0.0000}   \\
    StableDiffusion v2 ~\cite{podell2023sdxl}      & {0.5170}  & {0.2683} & {0.1010}    & {0.0437}   & {0.0120}   & {0.0040}    & {0.0010}   \\
    StableDiffusion XL Turbo~\cite{sauer2024adversarial} & 0.5886  & 0.3073  & 0.1343  & 0.0493    & 0.0190   & 0.0103 & 0.0003    \\
    PixArt-$\alpha$~\cite{chen2023pixart}                        & 0.6170 & 0.3133 & 0.1353 & 0.0490 & 0.0207 & 0.0097 & 0.0013   \\
    % DeepFloyd IF XL~\cite{deepfloyd2023if}   & - & -  & -  & -  & - & - & -   \\
    StableDiffusion XL~\cite{podell2023sdxl}     & {0.6237}  & {0.3517}  & {0.1670}  & {0.0553}    & {0.0273}   & {0.0127}    & {0.0050}   \\
    Playground V2.5~\cite{li2024playground} & 0.6963  & 0.3830 & 0.1917 & 0.0690 & 0.0373 &  0.0113 &  0.0017 \\
    StableDiffusion 3.5~\cite{esser2024scaling} & 0.7003  & 0.4467  & 0.3010 & 0.1673    & 0.1487 & 0.0597 & 0.0430   \\
    FLUX.1-dev~\cite{flux2024} & 0.6647  & 0.4127  & 0.2680 & 0.1620 & 0.1173 & 0.0567 & 0.0372   \\
    IterComp~\cite{zhang2024itercomp} & 0.6570& 0.3790 & 0.2143 & 0.0740 & 0.0417 & 0.0213 & 0.0040 \\
    FlowGRPO~\cite{liu2025flow} & 0.7600  & 0.5457  & 0.3517  & 0.2103 & 0.1727  &  0.0827 & 0.0713   \\
    PrefGRPO~\cite{Pref-GRPO&UniGenBench} & 0.7037 & 0.4790 & 0.3023 & 0.2120 & 0.1603 & 0.0897 & 0.0647  \\
    Qwen-Image~\cite{wu2025qwen} & \underline{0.8183} & \underline{0.6583} & \underline{0.5060} & \underline{0.3803} & \underline{0.3353} & \underline{0.2213} & \underline{0.1747} \\

    \cellcolor{cyan!10}\ours & \cellcolor{cyan!10}\textbf{0.8410} & \cellcolor{cyan!10}\textbf{0.7063} & \cellcolor{cyan!10}\textbf{0.5700} & \cellcolor{cyan!10}\textbf{0.3947} & \cellcolor{cyan!10}\textbf{0.3560} & \cellcolor{cyan!10}\textbf{0.2317} & \cellcolor{cyan!10}\textbf{0.1883} \\

    \bottomrule
    \end{tabular}
    }
    % \vspace{-2em}
    \end{subtable}
    \label{tab:full_mark}
% \end{table}
    \nextfloat 
% \begin{table}[ht]
    % \vspace{-3em}
    \centering
    \small
    \setlength\arrayrulewidth{1pt}
    \setlength\tabcolsep{6pt}
    \caption{\textbf{Concept Fraction Score of T2I Models on \conceptmix.} We report the concept fraction score across difficulty levels $k=1$ to $k=7$, where $k$ denotes the number of required concepts. As $k$ increases, the scores of all models decrease, but at different rates. \textbf{Bold} and \underline{underline} indicate the best and second-best results, respectively.
    }
    \begin{subtable}[t]{\textwidth}
    \centering
    \resizebox{\linewidth}{!}{
    \begin{tabular}{lccccccc}
    \toprule
     Models & $k=1$   & $k=2$ & $k=3$ & $k=4$   & $k=5$ & $k=6$ & $k=7$ \\ 
    \midrule 
    \rowcolor{gray!15}\multicolumn{8}{c}{\textit{Autoregressive Models}} \\
    \midrule 
    Janus~\cite{wu2025janus}  & 0.7385 & 0.6688 & 0.6318 & 0.5681 & 0.5602 & 0.5585 & 0.5285 \\
    Show-o~\cite{xie2024show} & 0.8435 & 0.7663 & 0.7260 & 0.6741 & 0.6691 & 0.6378 & 0.6331 \\
    HermesFlow~\cite{yang2025hermesflow} & 0.8362 & 0.7668 & 0.7193 & 0.6775 & 0.6705 & 0.6390 & 0.6348 \\
    \midrule 
    \rowcolor{gray!15}\multicolumn{8}{c}{\textit{Non-Autogressive Models}} \\
    \midrule
    Stable Diffusion v1.4\cite{rombach2022high} & 0.7153 & 0.5897 & 0.5190 & 0.4508  & 0.4105  & 0.3807  & 0.3475  \\
    Stable Diffusion v2~\cite{rombach2022high} & 0.7397 & 0.6467 & 0.5822 & 0.5086 & 0.4702 & 0.4458  & 0.4228  \\
    StableDiffusion XL Turbo~\cite{sauer2024adversarial} & 0.7824 & 0.6782 & 0.6324 & 0.5661 & 0.5330  & 0.4968  & 0.4630  \\
    PixArt-$\alpha$~\cite{chen2023pixart} &  0.7900 & 0.6858 & 0.6312 & 0.5634 & 0.5262  & 0.4936 & 0.4625  \\
    StableDiffusion XL~\cite{podell2023sdxl} & 0.8043 & 0.7169 & 0.6657 & 0.5958 & 0.5727  & 0.5337  & 0.5023  \\
    % DeepFloyd IF XL~\cite{deepfloyd2023if} & - & -& - & -  & -  & - & - \\
    Playground V2.5~\cite{li2024playground} & 0.8399 & 0.7298 & 0.6701 & 0.5962  & 0.5800 & 0.5452 & 0.5000 \\
    StableDiffusion 3.5~\cite{esser2024scaling} & 0.8367  & 0.7642 & 0.7368 & 0.6937    &  0.6943 &  0.6603 & 0.6565   \\
    FLUX.1-dev~\cite{flux2024} & 0.8162  & 0.7430 & 0.7093 & 0.6706  & 0.6779 & 0.6467 & 0.6576   \\
    IterComp~\cite{zhang2024itercomp} & 0.8247 & 0.7293 & 0.6984 & 0.6139 & 0.6038 & 0.5727 & 0.5419 \\
    FlowGRPO~\cite{liu2025flow} & 0.8747 & 0.8144  & 0.7726  & 0.7235    & {0.7263}  & 0.7010  & 0.6953  \\
    Pref-GRPO~\cite{Pref-GRPO&UniGenBench} & 0.8395 & 0.7819 & 0.7394 & 0.7159 & 0.7129 & 0.6868 & 0.6846 \\
    Qwen-Image~\cite{wu2025qwen} & \underline{0.9052} & \underline{0.8600} & \underline{0.8403} & \textbf{0.8193} & \underline{0.8171} & \underline{0.7869} & \textbf{0.7942} \\
    \cellcolor{cyan!10}\ours & \cellcolor{cyan!10}\textbf{0.9147}  & \cellcolor{cyan!10}\textbf{0.8850} & \cellcolor{cyan!10}\textbf{0.8635} & \cellcolor{cyan!10}{\underline{0.8188}}  & \cellcolor{cyan!10}\textbf{0.8214} & \cellcolor{cyan!10}\textbf{0.7917} & \cellcolor{cyan!10}{\underline{0.7884}}   \\
    \bottomrule
    \end{tabular}%
    }
    % \caption{\textbf{Concept fraction score:} average proportion of satisfied visual concepts. }
    \end{subtable}
    % \vspace{-1em}
\label{tab:concept_fraction}
\end{table}

\subsection{Experimental Results}
\label{sec:main_results}
\noindent\textbf{Results on \conceptmix.}
We first evaluate our proposed method on the \conceptmix benchmark, which rigorously tests a model's ability to handle mixed-concept prompts without attribute leakage. We report the Full Mark Score (Table~\ref{tab:main}) and Concept Fraction Score (Table~\ref{tab:concept_fraction}) across varying difficulty levels ($k \in [1, 7]$), where $k$ represents the number of distinct visual concepts required.

\noindent\textbf{Full Mark Score of T2I Models.}
\label{sec:full_mark}
As shown in Table~\ref{tab:full_mark}, \ours~consistently outperforms all evaluated baselines across every difficulty level. While the performance of existing models drops sharply as $k$ increases due to attribute leakage and object omission, \ours~ demonstrates remarkable robustness, achieving the highest score of 0.8410 at $k=1$. Most notably, at the extreme complexity of $k=7$, our model successfully maintains a score of 0.1883. This not only yields a substantial improvement over the base FLUX.1-dev model (0.0372) but also decisively surpasses the highly competitive Qwen-Image (0.1747). These results empirically prove that our approach effectively handles dense multi-concept prompts, setting a new state-of-the-art for strict multi-concept generation.

\noindent\textbf{Concept Fraction Score of T2I Models.}
\label{sec:concept_fraction}
As shown in Table~\ref{tab:concept_fraction}, we also report the Concept Fraction Score to evaluate the average proportion of correctly generated concepts within a prompt, accounting for partial successes. Similar to the strict exact-match evaluation, our proposed method consistently demonstrates superior performance across all difficulty levels. Notably, even under the extreme compositional constraint of $k=7$, our model successfully preserves a high concept fraction score of 0.7884. This result not only significantly improves upon the base FLUX.1-dev (0.6576) and the RL-aligned Pref-GRPO (0.6846), but also closely rivals or exceeds the massive Qwen-Image model across most complexity levels. These findings confirm that difficulty-aware compositional reward effectively prevents catastrophic object omission, enabling the model to retain strong attribute binding even when densely packed with concepts.

\begin{wraptable}{r}{0.4\linewidth}
\vspace{-35pt}
\centering
\caption{\textbf{Soft-TIFA Evaluation of T2I Models on GenEval~2.} 
}
\small
\setlength{\tabcolsep}{4pt}
\renewcommand{\arraystretch}{0.95}

\resizebox{\linewidth}{!}{
\begin{tabular}{lcc}
\toprule
\multirow{2}{*}{\textbf{Model}} & \textbf{Soft} & \textbf{Soft} \\
& \textbf{TIFA$_\text{AM}$} & \textbf{TIFA$_\text{GM}$} \\
\midrule

\rowcolor{gray!15}\multicolumn{3}{c}{\textit{Autoregressive Models}} \\
\midrule
Janus~\cite{wu2025janus} & 52.9 & 11.0 \\
Show-o~\cite{xie2024show} & 63.4 & 16.0 \\
HermesFlow~\cite{yang2025hermesflow} & 63.5 & 16.1 \\
\midrule

\rowcolor{gray!15}\multicolumn{3}{c}{\textit{Non-Autogressive Models}} \\
\midrule
SD 2.1~\cite{rombach2022high} & 40.8 & 5.2 \\
SDXL~\cite{podell2023sdxl} & 50.1 & 9.1 \\
SD3.5-M~\cite{esser2024scaling} & 63.7 & 15.8 \\
FLUX.1-dev~\cite{flux2024} & 64.1 & 17.1 \\
IterComp~\cite{zhang2024itercomp} & 53.0 & 8.5 \\
FlowGRPO~\cite{liu2025flow} & 70.9 & 21.8 \\
PrefGRPO~\cite{Pref-GRPO&UniGenBench} & 70.3 & 23.0 \\
Qwen-Image~\cite{wu2025qwen} & \textbf{80.0} & 31.4 \\
\cellcolor{cyan!10}\ours & \cellcolor{cyan!10}\textbf{80.0} & \cellcolor{cyan!10}\textbf{34.0} \\
\bottomrule

\end{tabular}
}
\label{tab:GenEval 2}
\vspace{-25pt}
\end{wraptable}

\noindent\textbf{Evaluation on GenEval~2.}
\label{sec:soft_tifa}
Table~\ref{tab:GenEval 2} presents the results on GenEval 2 using the Soft-TIFA metric, which evaluates atom-level ($\text{TIFA}_\text{AM}$) and prompt-level ($\text{TIFA}_\text{GM}$) correctness. Our method demonstrates superior generalization by establishing a new state-of-the-art, achieving $\textbf{80.0}$ in $\text{TIFA}_\text{AM}$ and $\textbf{34.0}$ in $\text{TIFA}_\text{GM}$. This decisively outperforms existing RL-aligned baselines (FlowGRPO, PrefGRPO) and notably surpasses the large-scale Qwen-Image in prompt-level correctness (34.0 vs. 31.4). These results confirm that our fine-grained reward structure successfully enforces complex relational reasoning and holistic composition, generalizing robustly to unseen prompt distributions rather than merely relying on memorized attribute associations.

\noindent\textbf{Evaluation on T2I-CompBench.}
\label{sec:compbench}
To systematically evaluate fine-grained compositional capabilities, we present the results on T2I-CompBench in Table~\ref{tab:t2icomp}. This benchmark assesses models across multiple dimensions, including attribute binding (color, shape, texture), object relationships (spatial, non-spatial), numeracy, and complex compositions. Our proposed framework achieves state-of-the-art results in the majority of individual categories and consistently yields the highest overall averages. Specifically, our method equipped with Stable Diffusion 3.5 records a state-of-the-art Avg score of $\textbf{0.6141}$, excelling significantly in fundamental attribute binding (Color: $\textbf{0.8692}$, Shape: $\textbf{0.6427}$, Texture: $\textbf{0.7959}$) and spatial relationships ($\textbf{0.5475}$). Furthermore, when applied to the FLUX.1-dev backbone, our method demonstrates exceptional proficiency in higher-order reasoning, achieving the highest score in the Complex category ($\textbf{0.4909}$) and an impressive Avg of $\underline{0.5961}$. These comprehensive improvements across diverse and granular criteria confirm that our difficulty-aware reward optimization avoids overfitting to specific attributes, ensuring a well-balanced and robust compositional generation capability.

\begin{table}[t!]
% % \vspace{-1.5em}
\centering
\caption{\textbf{Evaluation Results on T2I-CompBench \cite{huang2023t2i}. }
We report attribute binding, object relationships, numeracy, complex compositions, and their overall \textbf{Avg} (arithmetic mean over the seven metrics). 
\textbf{Bold} and \underline{underline} indicate the best and second-best results, respectively.} 
% \vspace{-2mm}
\label{benchmark:t2icompbench}
\resizebox{1\linewidth}{!}{ 
\begin{tabular}
{lcccccccc}
\toprule
\multicolumn{1}{c}
{\multirow{2}{*}{\bf Model}} & \multicolumn{3}{c}{\bf Attribute Binding } & \multicolumn{2}{c}{\bf Object Relationship} & \multirow{2}{*}{\bf Numeracy$\uparrow$}& \multirow{2}{*}{\bf Complex$\uparrow$} & \multirow{2}{*}{\bf Avg$\uparrow$}
\\
\cmidrule(lr){2-4}\cmidrule(lr){5-6}

&
{\bf Color $\uparrow$ } &
{\bf Shape$\uparrow$} &
{\bf Texture$\uparrow$} &
{\bf Spatial$\uparrow$} &
{\bf Non-Spatial$\uparrow$} &
&
&
\\
\cmidrule(lr){2-9}
   & BLIP-VQA &  BLIP-VQA &  BLIP-VQA & UniDet & CLIPscore & UniDet & 3-in-1 &  \\

\midrule
% DALL-E 2 \cite{ramesh2022hierarchical}  & 0.5750 & 0.5464 & 0.6374 & 0.1283 & 0.3043 & 0.4873 & 0.3696 & 0.4355 \\
Stable Diffusion XL \cite{podell2023sdxl} & 0.5879 & 0.4687 & 0.5299 & 0.2133 & 0.3119 & 0.4991 & 0.3237 & 0.4192 \\
PixArt-$\alpha$ \cite{chen2023pixart} & {0.4123} & {0.3809} &  {0.4559} &  {0.2057} &  {0.3083} &{0.5085} &  {0.3604} & 0.3760 \\
DALL-E 3 \cite{betker2023improving} & 0.7785 & {0.6205} & 0.7036 & 0.2865 & 0.3003 & 0.5926 & 0.3373 & 0.5170 \\
Stable Diffusion 3.5 \cite{esser2024scaling} & {0.7926} & {0.5617} & {0.7338} & {0.2914} & {0.3139} & 0.5898 & {0.3842} & 0.5239 \\
FLUX.1-dev~\cite{flux2024} & {0.7358} & {0.4802} & {0.5989} & {0.2461} & {0.3067} & 0.6107 & {0.4281} & 0.4866 \\
IterComp \cite{zhang2024itercomp} & 0.7295 & {0.5356} & 0.6526 & 0.2471 & \textbf{0.3227} & 0.5430 & 0.3883 & 0.4884 \\
FlowGRPO\cite{liu2025flow} & {0.8263} & {0.6177} & {0.7241} & \underline{0.5237} & {0.3184} & {0.6987} & {0.3905} & 0.5856 \\
PrefGRPO~\cite{Pref-GRPO&UniGenBench} & {0.7895} & {0.5595} & {0.6688} & {0.2846} & {0.3102} & 0.6559 & \underline{0.4661} & 0.5335 \\
{NanoBanana}~\cite{comanici2025gemini} &{0.8194} & \underline{0.6275} &  {0.7159} & {0.3240} & {0.3023}  & {0.6321} & {0.3741} & 0.5422 \\
Qwen-Image\cite{wu2025qwen} &{0.8395} & {0.5882} &  \underline{0.7407} & {0.4454} & {0.3136}  & \textbf{0.7553} & {0.4414} & 0.5892 \\
\cellcolor{cyan!10}StableDiffusion 3.5 + \ours  & \cellcolor{cyan!10}\textbf{0.8692} & \cellcolor{cyan!10}\textbf{0.6427} & \cellcolor{cyan!10}\textbf{0.7959} & \cellcolor{cyan!10}\textbf{0.5475} & \cellcolor{cyan!10}\underline{0.3201} & \cellcolor{cyan!10}\underline{0.7200} & \cellcolor{cyan!10}{0.4036} & \cellcolor{cyan!10}\textbf{0.6141}  \\

\cellcolor{cyan!10}FLUX.1-dev + \ours  & \cellcolor{cyan!10}\underline{0.8607} & \cellcolor{cyan!10}{0.6188} & \cellcolor{cyan!10}{0.7206} & \cellcolor{cyan!10}{0.4577} & \cellcolor{cyan!10}{0.3171} & \cellcolor{cyan!10}{0.7066} & \cellcolor{cyan!10}\textbf{0.4909} & \cellcolor{cyan!10}\underline{0.5961}  \\

\bottomrule
\end{tabular}
}

\label{tab:t2icomp}
\vspace{0em}
\end{table}

\begin{table}[t]
    \centering
    \caption{\textbf{Ablation on Design Choices.} We explicitly isolate the impact of our baseline optimization, training data, and correlation mechanism. \textbf{DRO} (Decoupled Reward Optimization) denotes our base multi-reward setup using MixGRPO~\cite{li2025mixgrpo} and GDPO~\cite{liu2026gdpo}. \textbf{Single-Attr.} and \textbf{Multi-Concept} represent the inclusion of training sets consisting of single-attribute binding and multi concepts, respectively. \textbf{CR} denotes Correlation-based Reweighting.}

    \label{tab:ablation}
    \setlength{\tabcolsep}{10pt} % 헤더가 1줄로 통합되어 여백을 넉넉히 주었습니다
    \renewcommand{\arraystretch}{1.1}
    \resizebox{\linewidth}{!}{
    \begin{tabular}{l cccc cc}
    \toprule
    \textbf{Method} & \textbf{DRO} & \textbf{Single-Attr.} & \textbf{Multi-Concept} & \textbf{CR} & \textbf{Avg. Full Mark} & \textbf{Avg. Concept Frac.} \\
    % 전처리부에 \usepackage{multirow} 이 포함되어 있어야 합니다.
    
    \midrule
    SD3.5-M (Base) & - & - & - & - & 0.2667 & 0.7203 \\
    \midrule
    \multirow{4}{*}{Ours} & $\checkmark$ & $\times$ & $\times$ & $\times$ & 0.2713 & 0.7411 \\
                          & $\checkmark$ & $\checkmark$ & $\times$ & $\times$ & 0.3305 & 0.7752 \\
                          & $\checkmark$ & $\checkmark$ & $\checkmark$ & $\times$ & 0.3531 & 0.7993 \\
                          & $\checkmark$ & $\checkmark$ & $\checkmark$ & $\checkmark$ & \textbf{0.3913} & \textbf{0.8126} \\
    \bottomrule
    \end{tabular}
    }

\end{table}

\subsection{Ablation Study}
\label{sec:ablation}
To provide a more in-depth discussion of the findings from Table 5 in the main manuscript, Table~\ref{tab:ablation} delves deeper by explicitly isolating the contributions of our reward optimization baseline (\textbf{DRO}), training datasets (\textbf{Single-Attr.} and \textbf{Multi-Concept}), and our core contribution (\textbf{CR}). To comprehensively analyze the optimization, we evaluate both Avg. Full Mark score and Avg. Concept Fraction score. 

The results clearly demonstrate the necessity of each proposed component. Applying our advanced optimization (\textbf{DRO}) solely on the standard baseline dataset severely degrades performance (Avg. Full Mark score: 0.2713), proving that standard datasets lack the structural density required to ground fine-grained multi-concept rewards. Swapping to our proposed \textbf{Single-Attr.} and \textbf{Multi-Concept} datasets provides the essential training signals, stabilizing the optimization and recovering the performance to 0.3531. Ultimately, even with optimal data and baseline optimization, the breakthrough is strictly driven by \textbf{CR}. By explicitly resolving the remaining negative correlations among conflicting concepts, \textbf{CR} achieves the state-of-the-art performance (Avg. Full Mark score: 0.3913, Avg. Concept Fraction: 0.8126).

\subsection{Qualitative Result on ConceptMix}
\label{sec:qual}
\begin{figure}[t]
\centering
\includegraphics[width=\textwidth]{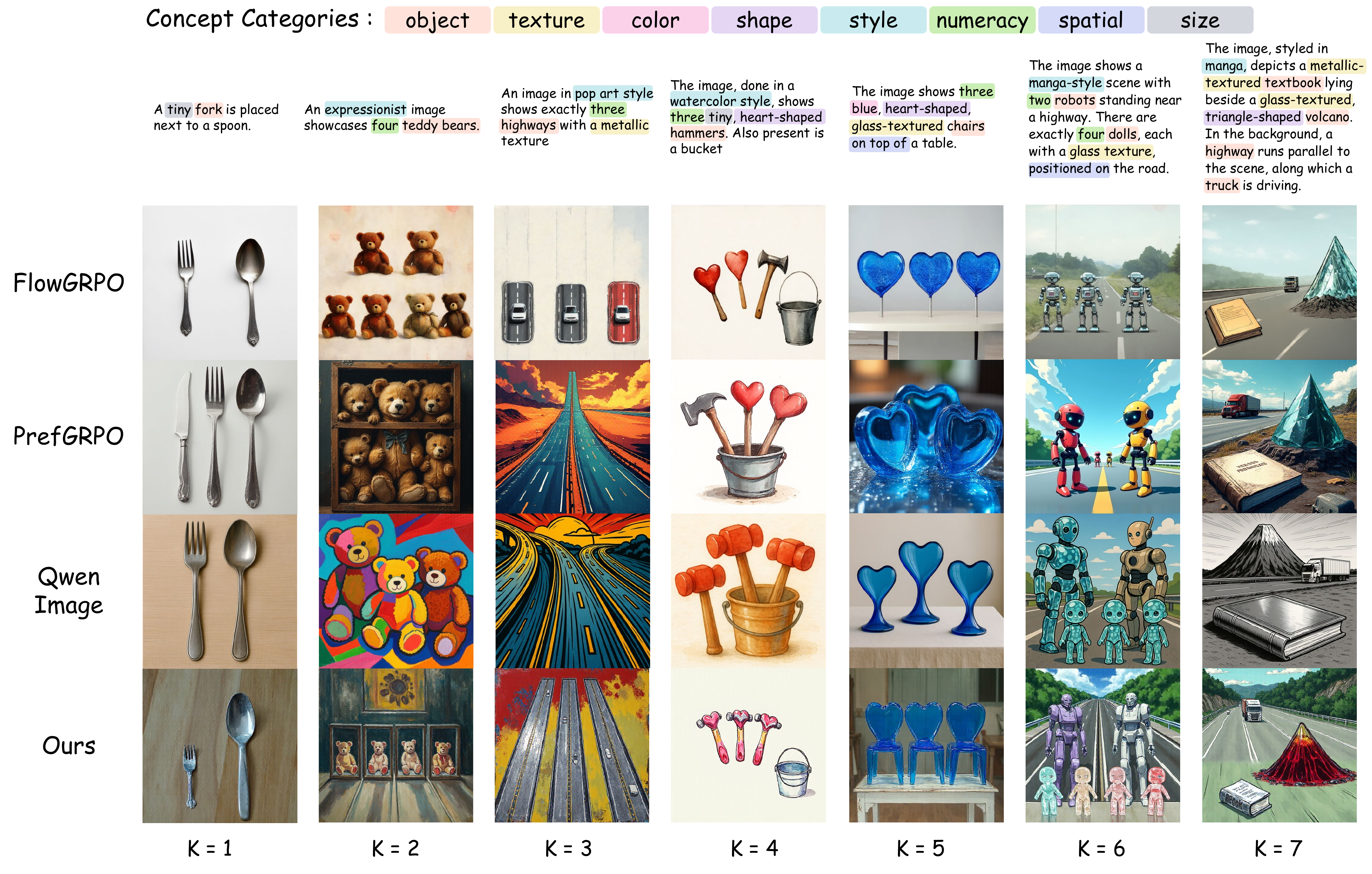} % 실제 파일 경로로 수정하세요
\vspace{-2em}
\caption{\textbf{Qualitative comparison across varying prompt complexity ($K=1 \sim 7$).} Baseline models frequently exhibit concept omission or attribute leakage as complexity increases. In contrast, our method consistently maintains high faithfulness and accurate attribute binding even under extreme constraints ($K=7$).}
\vspace{-2em}
\label{fig:qualitative}
\end{figure}

Figure~\ref{fig:qualitative} illustrates the qualitative performance across complexity levels $K=1$ to $K=7$. While baselines often suffer from object omission or misaligned attributes as $K$ increases, our method consistently generates faithful images by accurately aligning a diverse categories(\eg., object, color, numeracy, spatial, size).

\section{Discussions}
\label{sec:discussion}

The current design of CMO also suggests several directions for future work. First, future studies may develop more adaptive training data construction strategies that adjust the sampling frequency of each concept group based on observed failure patterns or compositional conflicts. In particular, the sampling ratio could be updated according to which reward groups are repeatedly involved in failed generations, rather than being fixed a priori. This would make the training prompt distribution itself part of the difficulty-aware optimization loop. Such data-centric refinement could help determine whether failures in specific attributes can be mitigated by exposing the model to more targeted concept combinations. Second, CMO may be extended to preference-aware settings by jointly considering compositional prompts and perceptual-quality constraints during data construction and reward design. Finally, studying the relationship between pre-training data scale and post-training multi-reward optimization would further clarify when structured reward optimization is most effective for compositional generation.

\section{Conclusion}
\label{sec:conclusion}

In this work, we address the persistent challenge of multi-concept compositional generation in text-to-image models,where models often struggle to align multiple concepts within a prompt simultaneously. To overcome this, we propose Correlation-Weighted Multi-Reward Optimization (\ours). Crucially, by computing the correlation among concept-wise reward signals across generated samples, our framework explicitly estimates concept generation difficulty. This allows the model to dynamically assign higher optimization weights to negatively correlated, hard-to-align concepts. Extensive experiments on state-of-the-art architectures, including Stable Diffusion 3.5 and FLUX.1-dev, demonstrate that \ours significantly mitigates negative correlations among multi-concepts even under extreme prompt complexities. By establishing new state-of-the-art performance on challenging benchmarks like ConceptMix, Geneval 2, and T2I-CompBench, our findings demonstrate that designing a structured, difficulty-aware multi-reward landscape is the definitive key to mastering true compositional generation.

% \clearpage\mbox{}Page \thepage\ of the manuscript.
% \clearpage\mbox{}Page \thepage\ of the manuscript.
% \clearpage\mbox{}Page \thepage\ of the manuscript.
% \clearpage\mbox{}Page \thepage\ of the manuscript.
% \clearpage\mbox{}Page \thepage\ of the manuscript. This is the last page.
% Now we have reached the maximum length of an ECCV \ECCVyear{} submission (excluding references and acknowledgements).
% References should start immediately after the main text, but can continue past p.\ 14 if needed. 
% \clearpage  % TODO FINAL: This \clearpage needs to be removed from both review and camera-ready versions.

\section*{Acknowledgements}
This research was supported by the Institute of Information \& Communications Technology Planning \& Evaluation (IITP) grant funded by the Korea government(MSIT) (No. RS-2019-II190079, Artificial Intelligence Graduate School Program(Korea University), 1\%; High-Performance Research AI Computing Infrastructure Support at the 2 PFLOPS Scale, 1\%), Culture, Sports and Tourism R\&D Program through the Korea Creative Content Agency grant funded by the Ministry of Culture, Sports and Tourism in 2024(Project Name: International Collaborative Research and Global Talent Development for the Development of Copyright Management and Protection Technologies for Generative AI, Project Number: RS-2024-00345025, 10\%), the National Research Foundation of Korea(NRF) grant funded by the Korea government(MSIT) (No. RS-2024-00341514, 28\%; No. RS-2025-02263628, 60\%)

\bibliographystyle{splncs04}
\bibliography{main}

@String(CVPR  = {IEEE Conf. Comput. Vis. Pattern Recog.})

@String(ICLR  = {Int. Conf. Learn. Represent.})

@String(TOG   = {ACM Trans. Graph.})

@String(CVPR  = {CVPR})

@String(ICLR  = {ICLR})

@String(TOG   = {ACM TOG})

@article{liu2025flow,
  title={Flow-grpo: Training flow matching models via online rl},
  author={Liu, Jie and Liu, Gongye and Liang, Jiajun and Li, Yangguang and Liu, Jiaheng and Wang, Xintao and Wan, Pengfei and Zhang, Di and Ouyang, Wanli},
  journal={arXiv preprint arXiv:2505.05470},
  year={2025}
}

@article{xue2025dancegrpo,
  title={DanceGRPO: Unleashing GRPO on Visual Generation},
  author={Xue, Zeyue and Wu, Jie and Gao, Yu and Kong, Fangyuan and Zhu, Lingting and Chen, Mengzhao and Liu, Zhiheng and Liu, Wei and Guo, Qiushan and Huang, Weilin and others},
  journal={arXiv preprint arXiv:2505.07818},
  year={2025}
}

@article{liu2026gdpo,
  title={GDPO: Group reward-Decoupled Normalization Policy Optimization for Multi-reward RL Optimization},
  author={Liu, Shih-Yang and Dong, Xin and Lu, Ximing and Diao, Shizhe and Belcak, Peter and Liu, Mingjie and Chen, Min-Hung and Yin, Hongxu and Wang, Yu-Chiang Frank and Cheng, Kwang-Ting and others},
  journal={arXiv preprint arXiv:2601.05242},
  year={2026}
}

@inproceedings{rombach2022high,
  title={High-resolution image synthesis with latent diffusion models},
  author={Rombach, Robin and Blattmann, Andreas and Lorenz, Dominik and Esser, Patrick and Ommer, Bj{\"o}rn},
  booktitle={Proceedings of the IEEE/CVF conference on computer vision and pattern recognition},
  pages={10684--10695},
  year={2022}
}

@article{podell2023sdxl,
  title={Sdxl: Improving latent diffusion models for high-resolution image synthesis},
  author={Podell, Dustin and English, Zion and Lacey, Kyle and Blattmann, Andreas and Dockhorn, Tim and M{\"u}ller, Jonas and Penna, Joe and Rombach, Robin},
  journal={arXiv preprint arXiv:2307.01952},
  year={2023}
}

@article{chen2023pixart,
  title={Pixart-$alpha$: Fast training of diffusion transformer for photorealistic text-to-image synthesis},
  author={Chen, Junsong and Yu, Jincheng and Ge, Chongjian and Yao, Lewei and Xie, Enze and Wu, Yue and Wang, Zhongdao and Kwok, James and Luo, Ping and Lu, Huchuan and others},
  journal={arXiv preprint arXiv:2310.00426},
  year={2023}
}

@inproceedings{sauer2024adversarial,
  title={Adversarial diffusion distillation},
  author={Sauer, Axel and Lorenz, Dominik and Blattmann, Andreas and Rombach, Robin},
  booktitle={European Conference on Computer Vision},
  pages={87--103},
  year={2024},
  organization={Springer}
}

@inproceedings{esser2024scaling,
  title={Scaling rectified flow transformers for high-resolution image synthesis},
  author={Esser, Patrick and Kulal, Sumith and Blattmann, Andreas and Entezari, Rahim and M{\"u}ller, Jonas and Saini, Harry and Levi, Yam and Lorenz, Dominik and Sauer, Axel and Boesel, Frederic and others},
  booktitle={Forty-first international conference on machine learning},
  year={2024}
}

@article{ghosh2023geneval,
  title={Geneval: An object-focused framework for evaluating text-to-image alignment},
  author={Ghosh, Dhruba and Hajishirzi, Hannaneh and Schmidt, Ludwig},
  journal={Advances in Neural Information Processing Systems},
  volume={36},
  pages={52132--52152},
  year={2023}
}

@article{huang2023t2i,
  title={T2i-compbench: A comprehensive benchmark for open-world compositional text-to-image generation},
  author={Huang, Kaiyi and Sun, Kaiyue and Xie, Enze and Li, Zhenguo and Liu, Xihui},
  journal={Advances in Neural Information Processing Systems},
  volume={36},
  pages={78723--78747},
  year={2023}
}

@inproceedings{liu2022compositional,
  title={Compositional visual generation with composable diffusion models},
  author={Liu, Nan and Li, Shuang and Du, Yilun and Torralba, Antonio and Tenenbaum, Joshua B},
  booktitle={European conference on computer vision},
  pages={423--439},
  year={2022},
  organization={Springer}
}

@article{feng2022training,
  title={Training-free structured diffusion guidance for compositional text-to-image synthesis},
  author={Feng, Weixi and He, Xuehai and Fu, Tsu-Jui and Jampani, Varun and Akula, Arjun and Narayana, Pradyumna and Basu, Sugato and Wang, Xin Eric and Wang, William Yang},
  journal={arXiv preprint arXiv:2212.05032},
  year={2022}
}

@article{chefer2023attend,
  title={Attend-and-excite: Attention-based semantic guidance for text-to-image diffusion models},
  author={Chefer, Hila and Alaluf, Yuval and Vinker, Yael and Wolf, Lior and Cohen-Or, Daniel},
  journal={ACM transactions on Graphics (TOG)},
  volume={42},
  number={4},
  pages={1--10},
  year={2023},
  publisher={ACM New York, NY, USA}
}

@article{hu2024token,
  title={Token merging for training-free semantic binding in text-to-image synthesis},
  author={Hu, Taihang and Li, Linxuan and van de Weijer, Joost and Gao, Hongcheng and Shahbaz Khan, Fahad and Yang, Jian and Cheng, Ming-Ming and Wang, Kai and Wang, Yaxing},
  journal={Advances in Neural Information Processing Systems},
  volume={37},
  pages={137646--137672},
  year={2024}
}

@misc{flux2024,
    author={Black Forest Labs},
    title={FLUX},
    year={2024},
    howpublished={\url{https://github.com/black-forest-labs/flux}},
}

@inproceedings{peebles2023scalable,
  title={Scalable diffusion models with transformers},
  author={Peebles, William and Xie, Saining},
  booktitle={Proceedings of the IEEE/CVF international conference on computer vision},
  pages={4195--4205},
  year={2023}
}

@article{wu2025qwen,
  title={Qwen-image technical report},
  author={Wu, Chenfei and Li, Jiahao and Zhou, Jingren and Lin, Junyang and Gao, Kaiyuan and Yan, Kun and Yin, Sheng-ming and Bai, Shuai and Xu, Xiao and Chen, Yilei and others},
  journal={arXiv preprint arXiv:2508.02324},
  year={2025}
}

@article{betker2023improving,
  title={Improving image generation with better captions},
  author={Betker, James and Goh, Gabriel and Jing, Li and Brooks, Tim and Wang, Jianfeng and Li, Linjie and Ouyang, Long and Zhuang, Juntang and Lee, Joyce and Guo, Yufei and others},
  journal={Computer Science. https://cdn. openai. com/papers/dall-e-3. pdf},
  volume={2},
  number={3},
  pages={8},
  year={2023}
}

@article{li2024playground,
  title={Playground v2. 5: Three insights towards enhancing aesthetic quality in text-to-image generation},
  author={Li, Daiqing and Kamko, Aleks and Akhgari, Ehsan and Sabet, Ali and Xu, Linmiao and Doshi, Suhail},
  journal={arXiv preprint arXiv:2402.17245},
  year={2024}
}

@article{kamath2025geneval,
  title={GenEval 2: Addressing Benchmark Drift in Text-to-Image Evaluation},
  author={Kamath, Amita and Chang, Kai-Wei and Krishna, Ranjay and Zettlemoyer, Luke and Hu, Yushi and Ghazvininejad, Marjan},
  journal={arXiv preprint arXiv:2512.16853},
  year={2025}
}

@article{wu2024conceptmix,
  title={Conceptmix: A compositional image generation benchmark with controllable difficulty},
  author={Wu, Xindi and Yu, Dingli and Huang, Yangsibo and Russakovsky, Olga and Arora, Sanjeev},
  journal={Advances in Neural Information Processing Systems},
  volume={37},
  pages={86004--86047},
  year={2024}
}

@inproceedings{li2024genai,
  title={Genai-bench: A holistic benchmark for compositional text-to-visual generation},
  author={Li, Baiqi and Lin, Zhiqiu and Pathak, Deepak and Li, Jiayao Emily and Xia, Xide and Neubig, Graham and Zhang, Pengchuan and Ramanan, Deva},
  booktitle={Synthetic Data for Computer Vision Workshop@ CVPR 2024},
  year={2024}
}

@article{wei2025tiif,
  title={TIIF-Bench: How Does Your T2I Model Follow Your Instructions?},
  author={Wei, Xinyu and Zhang, Jinrui and Wang, Zeqing and Wei, Hongyang and Guo, Zhen and Zhang, Lei},
  journal={arXiv preprint arXiv:2506.02161},
  year={2025}
}

@inproceedings{dat2025vsc,
  title={Vsc: Visual search compositional text-to-image diffusion model},
  author={Dat, Do Huu and Hyeon-Woo, Nam and Mao, Po-Yuan and Oh, Tae-Hyun},
  booktitle={Proceedings of the IEEE/CVF International Conference on Computer Vision},
  pages={19153--19162},
  year={2025}
}

@article{zhang2024realcompo,
  title={Realcompo: Dynamic equilibrium between realism and compositionality improves text-to-image diffusion models},
  author={Zhang, Xinchen and Yang, Ling and Cai, Yaqi and Yu, Zhaochen and Xie, Jiake and Tian, Ye and Xu, Minkai and Tang, Yong and Yang, Yujiu and Cui, Bin},
  journal={CoRR},
  year={2024}
}

@inproceedings{dahary2024yourself,
  title={Be yourself: Bounded attention for multi-subject text-to-image generation},
  author={Dahary, Omer and Patashnik, Or and Aberman, Kfir and Cohen-Or, Daniel},
  booktitle={European Conference on Computer Vision},
  pages={432--448},
  year={2024},
  organization={Springer}
}

@article{li2025mixgrpo,
  title={Mixgrpo: Unlocking flow-based grpo efficiency with mixed ode-sde},
  author={Li, Junzhe and Cui, Yutao and Huang, Tao and Ma, Yinping and Fan, Chun and Yang, Miles and Zhong, Zhao},
  journal={arXiv preprint arXiv:2507.21802},
  year={2025}
}

@article{shao2024deepseekmath,
  title={Deepseekmath: Pushing the limits of mathematical reasoning in open language models},
  author={Shao, Zhihong and Wang, Peiyi and Zhu, Qihao and Xu, Runxin and Song, Junxiao and Bi, Xiao and Zhang, Haowei and Zhang, Mingchuan and Li, YK and Wu, Yang and others},
  journal={arXiv preprint arXiv:2402.03300},
  year={2024}
}

@article{wang2025coefficients,
  title={Coefficients-Preserving Sampling for Reinforcement Learning with Flow Matching},
  author={Wang, Feng and Yu, Zihao},
  journal={arXiv preprint arXiv:2509.05952},
  year={2025}
}

@article{bai2025qwen3,
  title={Qwen3-vl technical report},
  author={Bai, Shuai and Cai, Yuxuan and Chen, Ruizhe and Chen, Keqin and Chen, Xionghui and Cheng, Zesen and Deng, Lianghao and Ding, Wei and Gao, Chang and Ge, Chunjiang and others},
  journal={arXiv preprint arXiv:2511.21631},
  year={2025}
}

@inproceedings{cherti2023reproducible,
  title={Reproducible scaling laws for contrastive language-image learning},
  author={Cherti, Mehdi and Beaumont, Romain and Wightman, Ross and Wortsman, Mitchell and Ilharco, Gabriel and Gordon, Cade and Schuhmann, Christoph and Schmidt, Ludwig and Jitsev, Jenia},
  booktitle={Proceedings of the IEEE/CVF Conference on Computer Vision and Pattern Recognition},
  pages={2818--2829},
  year={2023}
}

@article{carion2025sam,
  title={Sam 3: Segment anything with concepts},
  author={Carion, Nicolas and Gustafson, Laura and Hu, Yuan-Ting and Debnath, Shoubhik and Hu, Ronghang and Suris, Didac and Ryali, Chaitanya and Alwala, Kalyan Vasudev and Khedr, Haitham and Huang, Andrew and others},
  journal={arXiv preprint arXiv:2511.16719},
  year={2025}
}

@article{lin2025depth,
  title={Depth anything 3: Recovering the visual space from any views},
  author={Lin, Haotong and Chen, Sili and Liew, Junhao and Chen, Donny Y and Li, Zhenyu and Shi, Guang and Feng, Jiashi and Kang, Bingyi},
  journal={arXiv preprint arXiv:2511.10647},
  year={2025}
}

@article{zhang2024itercomp,
  title={Itercomp: Iterative composition-aware feedback learning from model gallery for text-to-image generation},
  author={Zhang, Xinchen and Yang, Ling and Li, Guohao and Cai, Yaqi and Xie, Jiake and Tang, Yong and Yang, Yujiu and Wang, Mengdi and Cui, Bin},
  journal={arXiv preprint arXiv:2410.07171},
  year={2024}
}

@inproceedings{wu2025janus,
  title={Janus: Decoupling visual encoding for unified multimodal understanding and generation},
  author={Wu, Chengyue and Chen, Xiaokang and Wu, Zhiyu and Ma, Yiyang and Liu, Xingchao and Pan, Zizheng and Liu, Wen and Xie, Zhenda and Yu, Xingkai and Ruan, Chong and others},
  booktitle={Proceedings of the Computer Vision and Pattern Recognition Conference},
  pages={12966--12977},
  year={2025}
}

@article{xie2024show,
  title={Show-o: One single transformer to unify multimodal understanding and generation},
  author={Xie, Jinheng and Mao, Weijia and Bai, Zechen and Zhang, David Junhao and Wang, Weihao and Lin, Kevin Qinghong and Gu, Yuchao and Chen, Zhijie and Yang, Zhenheng and Shou, Mike Zheng},
  journal={arXiv preprint arXiv:2408.12528},
  year={2024}
}

@article{yang2025hermesflow,
  title={Hermesflow: Seamlessly closing the gap in multimodal understanding and generation},
  author={Yang, Ling and Zhang, Xinchen and Tian, Ye and Shang, Chenming and Xu, Minghao and Zhang, Wentao and Cui, Bin},
  journal={arXiv preprint arXiv:2502.12148},
  year={2025}
}

@article{he2508tempflow,
  title={Tempflow-grpo: When timing matters for grpo in flow models, 2025},
  author={He, Xiaoxuan and Fu, Siming and Zhao, Yuke and Li, Wanli and Yang, Jian and Yin, Dacheng and Rao, Fengyun and Zhang, Bo},
  journal={URL https://arxiv. org/abs/2508.04324},
  year={2025}
}

@article{Pref-GRPO&UniGenBench,
  title={Pref-GRPO: Pairwise Preference Reward-based GRPO for Stable Text-to-Image Reinforcement Learning},
  author={Wang, Yibin and Li, Zhimin and Zang, Yuhang and Zhou, Yujie and Bu, Jiazi and Wang, Chunyu and Lu, Qinglin and Jin, Cheng and Wang, Jiaqi},
  journal={arXiv preprint arXiv:2508.20751},
  year={2025}
}

@article{comanici2025gemini,
  title={Gemini 2.5: Pushing the frontier with advanced reasoning, multimodality, long context, and next generation agentic capabilities},
  author={Comanici, Gheorghe and Bieber, Eric and Schaekermann, Mike and Pasupat, Ice and Sachdeva, Noveen and Dhillon, Inderjit and Blistein, Marcel and Ram, Ori and Zhang, Dan and Rosen, Evan and others},
  journal={arXiv preprint arXiv:2507.06261},
  year={2025}
}

@article{ddpo,
  title={Training diffusion models with reinforcement learning},
  author={Black, Kevin and Janner, Michael and Du, Yilun and Kostrikov, Ilya and Levine, Sergey},
  journal={arXiv preprint arXiv:2305.13301},
  year={2023}
}

@article{dpok,
  title={Dpok: Reinforcement learning for fine-tuning text-to-image diffusion models},
  author={Fan, Ying and Watkins, Olivia and Du, Yuqing and Liu, Hao and Ryu, Moonkyung and Boutilier, Craig and Abbeel, Pieter and Ghavamzadeh, Mohammad and Lee, Kangwook and Lee, Kimin},
  journal={Advances in Neural Information Processing Systems},
  volume={36},
  pages={79858--79885},
  year={2023}
}

@article{yang2025qwen3,
  title={Qwen3 technical report},
  author={Yang, An and Li, Anfeng and Yang, Baosong and Zhang, Beichen and Hui, Binyuan and Zheng, Bo and Yu, Bowen and Gao, Chang and Huang, Chengen and Lv, Chenxu and others},
  journal={arXiv preprint arXiv:2505.09388},
  year={2025}
}

@inproceedings{wallace2024diffusion,
  title={Diffusion model alignment using direct preference optimization},
  author={Wallace, Bram and Dang, Meihua and Rafailov, Rafael and Zhou, Linqi and Lou, Aaron and Purushwalkam, Senthil and Ermon, Stefano and Xiong, Caiming and Joty, Shafiq and Naik, Nikhil},
  booktitle={Proceedings of the IEEE/CVF Conference on Computer Vision and Pattern Recognition},
  pages={8228--8238},
  year={2024}
}

@article{clark2023directly,
  title={Directly fine-tuning diffusion models on differentiable rewards},
  author={Clark, Kevin and Vicol, Paul and Swersky, Kevin and Fleet, David J},
  journal={arXiv preprint arXiv:2309.17400},
  year={2023}
}

@article{prabhudesai2023aligning,
  title={Aligning text-to-image diffusion models with reward backpropagation},
  author={Prabhudesai, Mihir and Goyal, Anirudh and Pathak, Deepak and Fragkiadaki, Katerina},
  year={2023}
}

@inproceedings{deng2024prdp,
  title={Prdp: Proximal reward difference prediction for large-scale reward finetuning of diffusion models},
  author={Deng, Fei and Wang, Qifei and Wei, Wei and Hou, Tingbo and Grundmann, Matthias},
  booktitle={Proceedings of the IEEE/CVF Conference on Computer Vision and Pattern Recognition},
  pages={7423--7433},
  year={2024}
}

@article{lu2024deepseek,
  title={Deepseek-vl: towards real-world vision-language understanding},
  author={Lu, Haoyu and Liu, Wen and Zhang, Bo and Wang, Bingxuan and Dong, Kai and Liu, Bo and Sun, Jingxiang and Ren, Tongzheng and Li, Zhuoshu and Yang, Hao and others},
  journal={arXiv preprint arXiv:2403.05525},
  year={2024}
}

@article{hu2022lora,
  title={Lora: Low-rank adaptation of large language models.},
  author={Hu, Edward J and Shen, Yelong and Wallis, Phillip and Allen-Zhu, Zeyuan and Li, Yuanzhi and Wang, Shean and Wang, Liang and Chen, Weizhu and others},
  journal={Iclr},
  volume={1},
  number={2},
  pages={3},
  year={2022}
}

@article{sedgwick2012pearson,
  title={Pearson’s correlation coefficient},
  author={Sedgwick, Philip},
  journal={Bmj},
  volume={345},
  year={2012},
  publisher={British Medical Journal Publishing Group}
}

@article{wang2025unified,
  title={Unified reward model for multimodal understanding and generation},
  author={Wang, Yibin and Zang, Yuhang and Li, Hao and Jin, Cheng and Wang, Jiaqi},
  journal={arXiv preprint arXiv:2503.05236},
  year={2025}
}

@misc{schuhmann2022laion,
  author = {Schuhmann, Christoph},
  title = {LAION Aesthetics},
  year = {2022},
}

@inproceedings{you2025teaching,
  title={Teaching large language models to regress accurate image quality scores using score distribution},
  author={You, Zhiyuan and Cai, Xin and Gu, Jinjin and Xue, Tianfan and Dong, Chao},
  booktitle={Proceedings of the Computer Vision and Pattern Recognition Conference},
  pages={14483--14494},
  year={2025}
}

@article{wu2023human,
  title={Human preference score v2: A solid benchmark for evaluating human preferences of text-to-image synthesis},
  author={Wu, Xiaoshi and Hao, Yiming and Sun, Keqiang and Chen, Yixiong and Zhu, Feng and Zhao, Rui and Li, Hongsheng},
  journal={arXiv preprint arXiv:2306.09341},
  year={2023}
}

@article{xu2023imagereward,
  title={Imagereward: Learning and evaluating human preferences for text-to-image generation},
  author={Xu, Jiazheng and Liu, Xiao and Wu, Yuchen and Tong, Yuxuan and Li, Qinkai and Ding, Ming and Tang, Jie and Dong, Yuxiao},
  journal={Advances in Neural Information Processing Systems},
  volume={36},
  pages={15903--15935},
  year={2023}
}

@article{saharia2022photorealistic,
  title={Photorealistic text-to-image diffusion models with deep language understanding},
  author={Saharia, Chitwan and Chan, William and Saxena, Saurabh and Li, Lala and Whang, Jay and Denton, Emily L and Ghasemipour, Kamyar and Gontijo Lopes, Raphael and Karagol Ayan, Burcu and Salimans, Tim and others},
  journal={Advances in neural information processing systems},
  volume={35},
  pages={36479--36494},
  year={2022}
}

@article{kirstain2023pick,
  title={Pick-a-pic: An open dataset of user preferences for text-to-image generation},
  author={Kirstain, Yuval and Polyak, Adam and Singer, Uriel and Matiana, Shahbuland and Penna, Joe and Levy, Omer},
  journal={Advances in neural information processing systems},
  volume={36},
  pages={36652--36663},
  year={2023}
}

@inproceedings{ma2025hpsv3,
  title={Hpsv3: Towards wide-spectrum human preference score},
  author={Ma, Yuhang and Wu, Xiaoshi and Sun, Keqiang and Li, Hongsheng},
  booktitle={Proceedings of the IEEE/CVF International Conference on Computer Vision},
  pages={15086--15095},
  year={2025}
}

@article{hurst2024gpt,
  title={Gpt-4o system card},
  author={Hurst, Aaron and Lerer, Adam and Goucher, Adam P and Perelman, Adam and Ramesh, Aditya and Clark, Aidan and Ostrow, AJ and Welihinda, Akila and Hayes, Alan and Radford, Alec and others},
  journal={arXiv preprint arXiv:2410.21276},
  year={2024}
}
\clearpage

\renewcommand{\thefigure}{\thesection.\arabic{figure}}
\renewcommand{\thetable}{\thesection.\arabic{table}}
\renewcommand{\theequation}{\thesection.\arabic{equation}}
\renewcommand{\thealgorithm}{\thesection.\arabic{algorithm}}

\counterwithin{figure}{section}
\counterwithin{table}{section}
\counterwithin{equation}{section}
\counterwithin{algorithm}{section}

\renewcommand{\footnoterule}{%
  \kern -3pt
  \hrule width 0.4\textwidth height 0.4pt
  \kern 2.6pt
}

\titlerunning{CMO}


\authorrunning{J.~Wi et al.}

\appendix
\renewcommand{\theHsection}{appendix.\Alph{section}}
\renewcommand{\theHsubsection}{appendix.\Alph{section}.\arabic{subsection}}

\vspace{0.5em}

\noindent The Appendix is organized as follows:

\vspace{0.5em}

\begin{center}
\renewcommand{\arraystretch}{1.35}
\begin{tabular}{@{}p{0.18\linewidth}p{0.74\linewidth}@{}}
\textbf{Section~\ref{sec:training_process}} &
describes the detailed training process of \ours, including the optimization pipeline (Algorithm~\ref{alg:cmo_training}), objective formulation, and hyperparameter configurations.
\\[0.35em]

\textbf{Section~\ref{sec:multi_reward}} &
provides detailed formulations and implementation specifics of the multi-concept reward functions, including spatial relations, attribute-level rewards, and size constraints.
\\[0.35em]

\textbf{Section~\ref{sec:image_quality}} &
reports quantitative evaluations of image quality and human preference alignment.
\\[0.35em]

\textbf{Section~\ref{sec:qualitative}} &
provides additional qualitative comparisons on compositional generation benchmarks.
\\[0.35em]
\textbf{Section~\ref{sec:cost}} &
analyzes the training cost and efficiency of our method compared to existing RL-based diffusion fine-tuning approaches.
\\[0.35em]

\textbf{Section~\ref{sec:addational_res}} &
presents additional experimental results on ConceptMix across different task complexity levels.
\\[0.35em]

\textbf{Section~\ref{sec:baselines_appendix}} &
describes the representative compositional T2I baselines used in our comparisons.
\end{tabular}
\end{center}

\clearpage

\section{Training Process of CMO}
\label{sec:training_process}
\noindent\textbf{CMO Training Pipeline.}
As outlined in Algorithm~\ref{alg:cmo_training}, we optimize the policy $\pi_\theta$ exclusively within the stochastic window $W(l)$ following the mixed ODE-SDE strategy of MixGRPO~\cite{li2025mixgrpo}. Here, $K$ denotes the total number of concepts, and the sliding window $W(l)$ is strictly defined as the initial 3 sampling steps. To ensure balanced multi-concept generation, 
\begin{algorithm}[ht!]
\caption{Correlation-Weighted Multi-Reward Optimization (CMO)}
\label{alg:cmo_training}
\begin{algorithmic}[1]
\Require Initial policy $\pi_\theta$, reward models $\{R_k\}_{k=1}^K$, prompt dataset $\mathcal{C}$, Total sampling steps $T$, group size $N$, sliding window $W(l)$, batch size $B$
\State Initialize left boundary of $W(l)$: $l \leftarrow 0$
\For{training iteration $m = 1, \dots, M$}
    \State Sample a batch of $B$ prompts $\mathcal{C}_B \sim \mathcal{C}$
    \State $\pi_{\theta_{\text{old}}} \leftarrow \pi_\theta$ \Comment{Update old policy model}
    \For{each prompt index $i = 1, \dots, B$ with prompt $\mathbf{c}^{(i)} \in \mathcal{C}_B$}
        
        \For{sample index $j = 1, \dots, N$} \Comment{Generate a group of images}
            \For{$t = 0, \dots, T - 1$}
                \If{$t \in W(l)$}
                    \State $\mathbf{x}_{t+1}^{(i,j)} \leftarrow \text{SDE\_Sampling}(\mathbf{x}_t^{(i,j)}, \mathbf{c}^{(i)})$
                \Else
                    \State $\mathbf{x}_{t+1}^{(i,j)} \leftarrow \text{ODE\_Sampling}(\mathbf{x}_t^{(i,j)}, \mathbf{c}^{(i)})$
                \EndIf
            \EndFor
        \EndFor
        
        \State Compute rewards $r_{k}^{(i,j)} = R_k(\mathbf{x}_T^{(i,j)}, \mathbf{c}^{(i)})$ for all $j, k$
        \State Compute difficulty $\alpha_k^{(i)}$ and weights $w_k^{(i)}$ from correlations of $\{r_k^{(i,j)}\}_{j=1}^N$
        
        \For{sample index $j = 1, \dots, N$}
            \State $A_k^{(i,j)} \leftarrow \displaystyle \frac{r_k^{(i,j)} - \mu_k^{(i)}}{\sigma_k^{(i)}}$ \Comment{Decoupled reward}
            \State $\hat{A}_{\text{total}}^{(i,j)} \leftarrow \displaystyle \frac{\sum_{k=1}^K w_k^{(i)} A_k^{(i,j)} - \mu_B}{\sigma_B + \epsilon}$ \Comment{Batch-normalized total advantage}
        \EndFor
        
        \For{$t \in W(l)$} \Comment{Policy optimization}
            \State $\theta \leftarrow \theta + \eta \nabla_\theta \mathcal{J}$ 
        \EndFor
        
    \EndFor
\EndFor
\end{algorithmic}
\end{algorithm}
we define the training objective $\mathcal{J}(\theta)$ as a clipped objective with KL regularization, replacing the standard advantage with our correlation-weighted multi-reward advantage, $\hat{A}_{\text{total}}^{(i,j)}$:
\begin{equation}
\begin{split}
\mathcal{J}(\theta) &= \mathbb{E}_{\mathbf{c}^{(i)} \sim \mathcal{C}, \{\mathbf{x}_T^{(i,j)}\}_{j=1}^N \sim \pi_{\theta_{\text{old}}}(\cdot | \mathbf{c}^{(i)})} \Bigg[ \frac{1}{N} \sum_{j=1}^N \frac{1}{|S|} \sum_{t \in S} \\
&\quad \bigg( \min \Big( r_t^{(i,j)}(\theta) \hat{A}_{\text{total}}^{(i,j)}, \text{clip}(r_t^{(i,j)}(\theta), 1-\varepsilon, 1+\varepsilon) \hat{A}_{\text{total}}^{(i,j)} \Big) - \beta \mathcal{J}_{\text{KL}} \bigg) \Bigg],
\end{split}
\end{equation}
where $\varepsilon$ is the clipping hyperparameter and $\beta$ controls the KL penalty. The policy ratio $r_t^{(i,j)}(\theta)$ between the current ($q_\theta$) and old ($q_{\theta_{\text{old}}}$) policies during the denoising step is:

\begin{equation}
r_t^{(i,j)}(\theta) = \frac{q_\theta(\mathbf{x}_{t+\Delta t} | \mathbf{x}_t, \mathbf{c}^{(i)})}{q_{\theta_{\text{old}}}(\mathbf{x}_{t+\Delta t} | \mathbf{x}_t, \mathbf{c}^{(i)})} ,
\end{equation}

To prevent the updated policy from deviating excessively from the reference model, we apply the KL divergence $\mathcal{J}_{\text{KL}}$. For diffusion models, this is analytically computed as the distance between their denoised predictions:
\begin{equation}
\mathcal{J}_{\text{KL}} = D_{\text{KL}}(\pi_\theta || \pi_{\theta_{\text{old}}}) = \frac{|| \mathbf{x}_{t+\Delta t}(\theta) - \mathbf{x}_{t+\Delta t}(\theta_{\text{old}}) ||^2}{2\sigma_t^2 \Delta t},
\end{equation}
Maximizing $\mathcal{J}(\theta)$ allows CMO to effectively prioritize hard-to-align concepts while maintaining generative stability.

\begin{table}[hb!]
\centering
\caption{Detailed training hyperparameters for CMO across different base models.}
\label{tab:hyperparameters}
\resizebox{0.45\linewidth}{!}{
\begin{tabular}{lcc}
\toprule
\textbf{Hyperparameter} & \textbf{SD3.5-M} & \textbf{FLUX.1-dev} \\
\midrule
Optimizer & \multicolumn{2}{c}{AdamW} \\
Learning Rate & \multicolumn{2}{c}{$3 \times 10^{-4}$} \\
Global Batch Size & \multicolumn{2}{c}{512} \\
Image Resolution & \multicolumn{2}{c}{$512 \times 512$} \\
Clipping Parameter ($\varepsilon$) & \multicolumn{2}{c}{0.1} \\
Stochastic SDE Steps ($|S|$) & \multicolumn{2}{c}{3} \\
Noise Level & \multicolumn{2}{c}{0.8} \\
KL Coefficient ($\beta$) & 0.015 & 0 \\
LoRA Rank ($r$) & 64 & 128 \\
Total Sampling Steps ($T$) & 10 & 6 \\
\bottomrule
\end{tabular}
}
\end{table}
\noindent\textbf{Training Configuration.} We provide the detailed training hyperparameters and configurations used for Correlation-Weighted Multi-Reward Optimization (CMO). The overall settings for both Stable Diffusion 3.5 and FLUX.1-dev are summarized in Table~\ref{tab:hyperparameters}.

\begin{figure}[ht!]
    \centering
    \includegraphics[width=\linewidth]{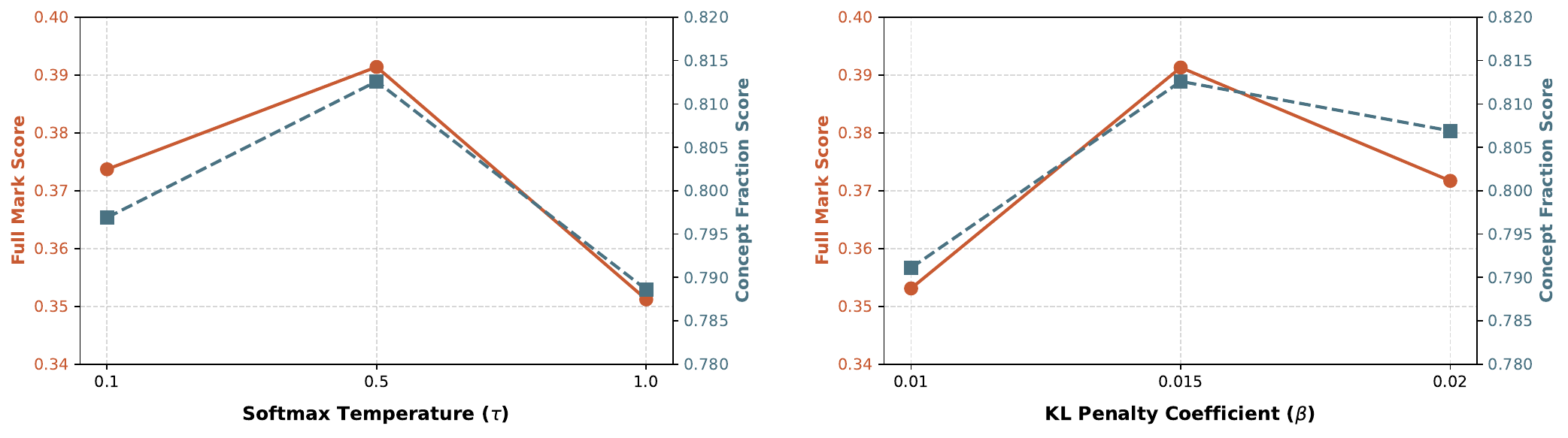}
    \caption{\textbf{Ablation on Hyperparameters $\tau$ and $\beta$.} We evaluate the impact of Softmax Temperature (Left) and KL Penalty Coefficient (Right) on compositional generation performance. Both hyperparameters exhibit a clear trade-off. Extreme values lead to failure to follow the fine-grained reward, while $\tau=0.5$ and $\beta=0.015$ strike the optimal balance for maximizing both the Full Mark and Concept Fraction scores.}
    \label{fig:hyperparam}
\end{figure}

\noindent\textbf{Impact of Hyperparameters ($\tau$ and $\beta$).}
Figure~\ref{fig:hyperparam} ablates the softmax temperature ($\tau$) in Eq.~10 and the KL penalty ($\beta$) using SD3.5-M. For $\tau$, the optimal value is 0.5; lower values ($\tau=0.1$) excessively sharpen weights, while higher values ($\tau=1.0$) overly smooth the distribution, negating the intended effect of prioritizing hard-to-align concepts. For the KL penalty, $\beta=0.015$ optimally balances reward optimization with prior preservation in SD3.5-M. A lower penalty ($\beta=0.01$) causes reward hacking, whereas a higher penalty ($\beta=0.02$) restricts the generative prior too strictly, hindering complex multi-concept alignments. Following MixGRPO~\cite{li2025mixgrpo}, we do not apply the KL penalty ($\beta=0$) when fine-tuning FLUX.1-dev. Thus, these hyperparameters are crucial for balancing optimization with stability.

\noindent\textbf{Impact of Group Size.} Table~\ref{tab:group_size} ablates the group size $N$.
\begin{wraptable}{r}{0.4\linewidth}
\vspace{-35pt} 
\centering
\caption{\textbf{Effect of Group Size.} We ablate the number of samples generated per prompt.}
\label{tab:group_size}
\footnotesize
\setlength{\tabcolsep}{4pt} 
\renewcommand{\arraystretch}{1.1}
\begin{tabular}{cc}
\toprule
\textbf{Size ($N$)} & \textbf{Avg. Full Mark} \\
\midrule
\textcolor{gray}{4} & \textcolor{gray}{0.0000 (Collapse)} \\
8 & 0.2934 \\
\rowcolor{cyan!10}
\textbf{16 (Default)} & \textbf{0.3919} \\
\bottomrule
\end{tabular}
\vspace{-25pt} 
\end{wraptable}
 A sufficient $N$ is strictly required, as our method relies on group-wise statistics for both advantage normalization and correlation estimation.
When $N=8$, the Avg. Full Mark score drops to 0.2934 due to noisier statistical estimates. At an extreme $N=4$, training completely collapses (0.0000), as such a minimal sample size severely compromises the reliability of group-wise statistical estimations, failing to provide stable optimization signals. While a larger $N$ could potentially yield more robust updates, scaling beyond $N=16$ incurs a prohibitive memory footprint. Thus, $N=16$ serves as an optimal configuration, balancing optimization stability with hardware constraints.

\section{Multi-Concept Reward for Training}
\label{sec:multi_reward}
\begin{figure}[t]
    \centering
    \includegraphics[width=\textwidth]{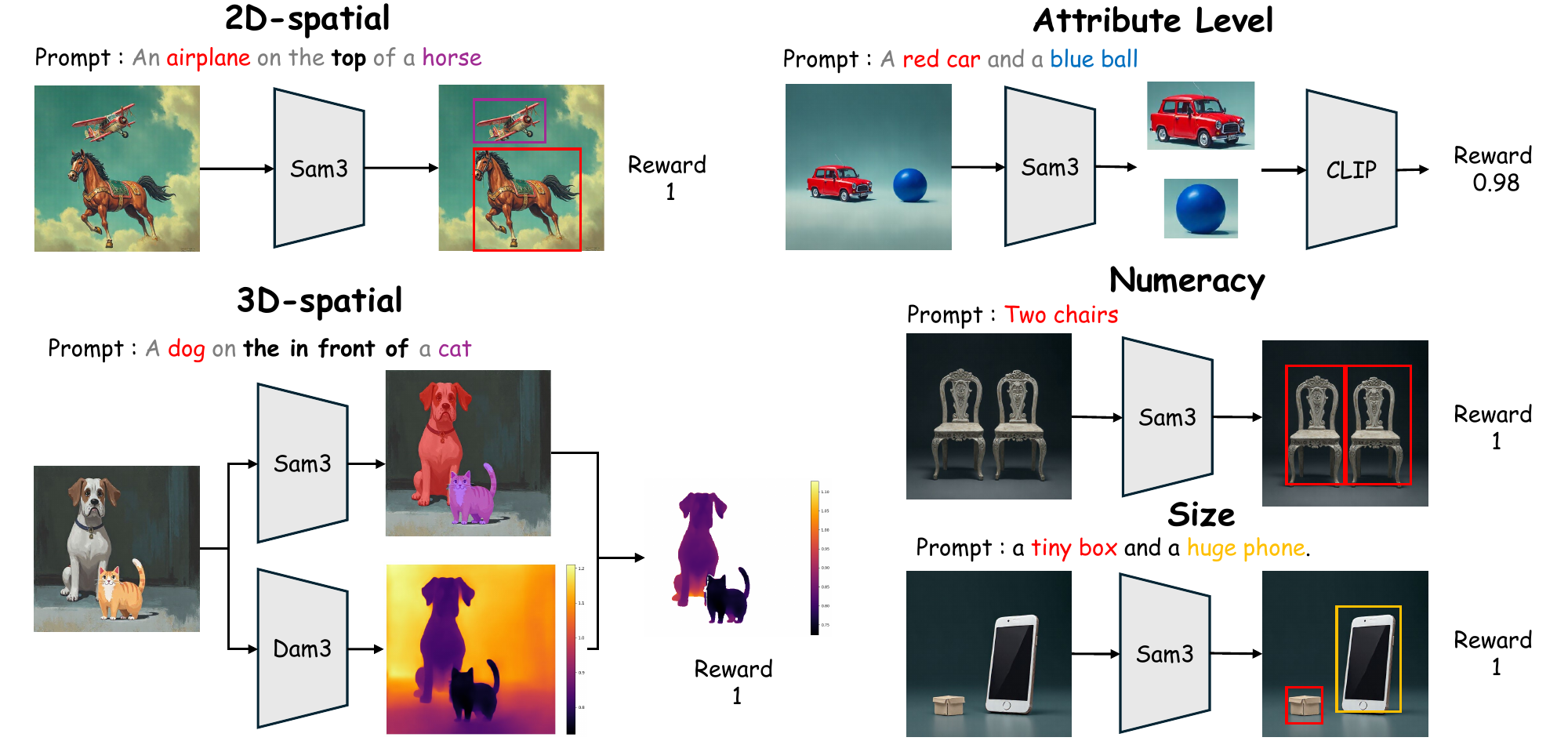}
    \caption{
        \textbf{Detailed Overview of Multi-Concept Reward.} 
        To effectively evaluate compositional generation, we decompose complex multi-concept prompts into five fine-grained categories: 2D-spatial, 3D-spatial, Attribute-level, Numeracy, and Size. Across all constraints, we first employ SAM~3 to achieve precise instance-level localization. Based on the extracted masks and bounding boxes, specific evaluators are routed: geometric and counting heuristics are applied for 2D-spatial, numeracy, and size constraints; Depth Anything~3 (Dam3) is utilized for relative depth comparison in 3D-spatial relations; and OpenCLIP is leveraged to compute the contrastive reward for visual attributes. This structured decomposition provides highly disentangled and accurate reward signals for our CMO framework.
    }
    \label{fig:reward_pipeline}
\end{figure}

We provide detailed formulations and implementation specifics for our multi-concept reward evaluation. As illustrated in Figure~\ref{fig:reward_pipeline}, to effectively evaluate and optimize compositional generation, we decompose complex multi-concept prompts into specific constraints. In the following subsections, we detail the exact reward functions for 2D and 3D spatial relations, the contrastive extension for attribute-level rewards, and the rank-based heuristic for size rewards used in our actual training pipeline.

\noindent\textbf{Spatial Relation Rewards.}
For 2D spatial relations, we apply strict boundary checks with a $10\%$ tolerance margin and axis alignment for a full reward of $1.0$, while falling back to center-point comparisons for a partial reward of $0.5$. 

Let $b_i = [x_{min}^i, y_{min}^i, x_{max}^i, y_{max}^i]$ and $b_j$ be the bounding boxes of objects $i$ and $j$, with center coordinates $(cx_i, cy_i)$ and $(cx_j, cy_j)$, respectively. We define the tolerance margins as $\tau_x = 0.1 \times \min(w_i, w_j)$ and $\tau_y = 0.1 \times \min(h_i, h_j)$. Indicator functions $\mathbb{I}_x, \mathbb{I}_y \in \{0, 1\}$ denote whether the two boxes intersect along the x-axis and y-axis. The reward functions are formulated as follows:

\paragraph{Horizontal Relations (Left / Right)}
\begin{equation}
r_{i,j}^{\text{left}} = \begin{cases} 1.0, & \text{if } x_{max}^i < x_{min}^j + \tau_x \text{ and } \mathbb{I}_y = 1 \\ 0.5, & \text{else if } cx_i < cx_j \\ 0.0, & \text{otherwise} \end{cases}
\end{equation}

\begin{equation}
r_{i,j}^{\text{right}} = \begin{cases} 1.0, & \text{if } x_{min}^i > x_{max}^j - \tau_x \text{ and } \mathbb{I}_y = 1 \\ 0.5, & \text{else if } cx_i > cx_j \\ 0.0, & \text{otherwise} \end{cases}
\end{equation}

\paragraph{Vertical Relations (Above / Below)}
\begin{equation}
r_{i,j}^{\text{above}} = \begin{cases} 1.0, & \text{if } y_{max}^i < y_{min}^j + \tau_y \text{ and } \mathbb{I}_x = 1 \\ 0.5, & \text{else if } cy_i < cy_j \\ 0.0, & \text{otherwise} \end{cases}
\end{equation}

\begin{equation}
r_{i,j}^{\text{below}} = \begin{cases} 1.0, & \text{if } y_{min}^i > y_{max}^j - \tau_y \text{ and } \mathbb{I}_x = 1 \\ 0.5, & \text{else if } cy_i > cy_j \\ 0.0, & \text{otherwise} \end{cases}
\end{equation}

\paragraph{Inclusion Relations (Inside / Outside)}
Let $A_i$ be the area of $B_i$, and $A_{i \cap j}$ be the intersection area.
\begin{equation}
r_{i,j}^{\text{inside}} = \begin{cases} 1.0, & \text{if } \frac{A_{i \cap j}}{A_i} \ge 0.95 \\ 0.0, & \text{otherwise} \end{cases}
\end{equation}

\begin{equation}
r_{i,j}^{\text{outside}} = \begin{cases} 1.0, & \text{if } \frac{A_{i \cap j}}{\min(A_i, A_j)} < 0.1 \\ 0.0, & \text{otherwise} \end{cases}
\end{equation}

For 3D spatial relations, we utilize the normalized depth map $\mathcal{D}(x)$ to compute the average depth $\bar{d}_i$ and $\bar{d}_j$ within the respective object masks. As defined in the main text, we apply a tolerance margin $\epsilon$ (empirically set to $0.02$) to account for depth estimation noise. The reward functions for depth-based relations are formulated as follows:
\begin{equation}
r_{i,j}^{\text{in front of}} = \begin{cases} 1.0, & \text{if } \bar{d}_i < \bar{d}_j - \epsilon \\ 0.5, & \text{if } \bar{d}_i < \bar{d}_j \\ 0.0, & \text{otherwise} \end{cases}
\end{equation}

\begin{equation}
r_{i,j}^{\text{behind}} = \begin{cases} 1.0, & \text{if } \bar{d}_i > \bar{d}_j + \epsilon \\ 0.5, & \text{if } \bar{d}_i > \bar{d}_j \\ 0.0, & \text{otherwise} \end{cases}
\end{equation}

\noindent\textbf{Contrastive Attribute-Level Reward.}
To provide additional details for the zero-shot classifier-based attribute reward in Eq.~(4), we describe how the candidate text feature $f_v$ is constructed and how the reward is computed over the candidate set. For each attribute category $k$ (e.g., color), we maintain a predefined candidate set $\mathcal{V}_k$ (e.g., \{red, green, blue, \dots\}). For each candidate attribute $v \in \mathcal{V}_k$, we construct multiple text prompts using predefined templates, such as ``a red object'' and ``an object with red color''. Let $\mathcal{T}(v)$ denote the set of template prompts for candidate attribute $v$. We encode each template prompt with the OpenCLIP text encoder $f_{\text{text}}(\cdot)$, average the resulting embeddings, and apply L2 normalization to obtain $f_v$:
\begin{equation}
f_v =
\frac{
\frac{1}{|\mathcal{T}(v)|}
\sum_{t \in \mathcal{T}(v)}
f_{\text{text}}(t)
}{
\left\|
\frac{1}{|\mathcal{T}(v)|}
\sum_{t \in \mathcal{T}(v)}
f_{\text{text}}(t)
\right\|_2
}.
\end{equation}

Given the cropped image feature $f_{\text{img}}(x \odot b_i)$ defined in the main manuscript, we compute the attribute reward by applying a softmax over all candidate attributes in $\mathcal{V}_k$. For the target attribute $v^* \in \mathcal{V}_k$ specified in the prompt, the reward is defined as:
\begin{equation}
\label{eq:attr_reward_supp}
r_i^{\text{attr}}
=
\frac{
\exp \left(
\langle f_{\text{img}}(x \odot b_i), f_{v^*} \rangle
\right)
}
{
\sum_{v \in \mathcal{V}_k}
\exp \left(
\langle f_{\text{img}}(x \odot b_i), f_v \rangle
\right)
}.
\end{equation}
This is the same formulation as Eq.~(4), with additional details on how the candidate text features are constructed. This formulation explicitly promotes the target attribute while suppressing competing attributes within the same category. For global attributes such as style, we use the feature of the entire generated image instead of a cropped object region.

\noindent\textbf{Size Reward.}
For size attributes (e.g., \textit{huge, tiny}), we evaluate the relative scale of the target object by comparing its area $A_i$ against the areas of all detected objects in the scene. We define $Rank(A_i)$ as the 1-based rank of the object's area (sorted in descending order for \textit{huge}, and ascending order for \textit{tiny}). The size reward decays quadratically as the rank deviates from the top position:
\begin{equation}
r_i^{\text{size}} = \frac{1}{Rank(A_i)^2}
\end{equation}

\section{Evaluation for Image Quality}
\label{sec:image_quality}

\begin{table*}[h]
\centering
\caption{\textbf{Quantitative Evaluation of Image Quality and Human Preference.} This experiment evaluates visual fidelity and alignment with human preferences on DrawBench~\cite{saharia2022photorealistic}. \ours successfully prevents the severe quality collapse observed in unregularized optimization (FlowGRPO w/o KL), achieving a highly competitive overall performance (AVG: 10.519) on par with FlowGRPO while strictly enforcing multi-concept.}

\resizebox{\textwidth}{!}{
\begin{tabular}{l cc ccccc c}
\toprule
\multirow{2}{*}{Model}  & \multicolumn{2}{c}{Image Quality} & \multicolumn{5}{c}{Preference Score} & \multirow{2}{*}{AVG$\uparrow$} \\
 \cmidrule(lr){2-3} \cmidrule(lr){4-8}
 & Aesthetic$\uparrow$ & DeQA$\uparrow$ & ImgRwd$\uparrow$ & PickScore$\uparrow$ & UniRwd$\uparrow$ & HPSv2$\uparrow$ & HPSv3$\uparrow$ & \\
\midrule

SD3.5-M~\cite{esser2024scaling} & 5.820 & 4.071 & 0.848 & 22.399 & 3.189 & 27.926 & 9.709 & 10.566 \\
\midrule
FlowGRPO (w/o KL)~\cite{liu2025flow} & 5.045 & 2.712 & 0.432 & 21.184 & 2.838 & 21.044 & 0.988 & 7.749 \\
FlowGRPO~\cite{liu2025flow}  & 5.766 & 4.019 & 1.014 & 22.575 & 3.508 & 27.133 & 9.628 & 10.520 \\
\ours & 5.653 & 3.829 & 1.182 & 22.330 & 3.461 & 27.283 & 9.896 & 10.519 \\
\bottomrule
\label{tab:preference}
\end{tabular}
} % resizebox 끝
\end{table*}
% 본문 삽입용 코드
As shown in Table~\ref{tab:preference}, we evaluate \ours and baselines across image quality (Aesthetic~\cite{schuhmann2022laion}, DeQA~\cite{you2025teaching}) and human preference benchmarks (ImageReward~\cite{xu2023imagereward}, PickScore~\cite{kirstain2023pick}, UnifiedReward~\cite{wang2025unified}, HPSv2~\cite{wu2023human}, and HPSv3~\cite{ma2025hpsv3}). For a robust evaluation, all images are generated using DrawBench~\cite{saharia2022photorealistic} test prompts with five different random seeds.

Overall, \ours achieves a highly competitive AVG score of $10.519$, matching the FlowGRPO baseline ($10.520$) and significantly outperforming the unregularized FlowGRPO w/o KL ($7.749$). Notably, our method successfully prevents the severe image quality collapse observed in unregularized reward optimization. While we observe a minor trade-off in base image quality compared to the original SD3.5-M, this is a natural consequence of strict compositional generation; accurately generating multi-concept inherently restricts the model from prioritizing aesthetic appeal over prompt alignment. Consequently, \ours reliably preserves visual fidelity while successfully aligning with human preferences.

\section{Qualitative Results for Compositional Generation.}
\label{sec:qualitative}
\begin{figure}[htbp]
    \centering
    \includegraphics[width=0.98\linewidth]{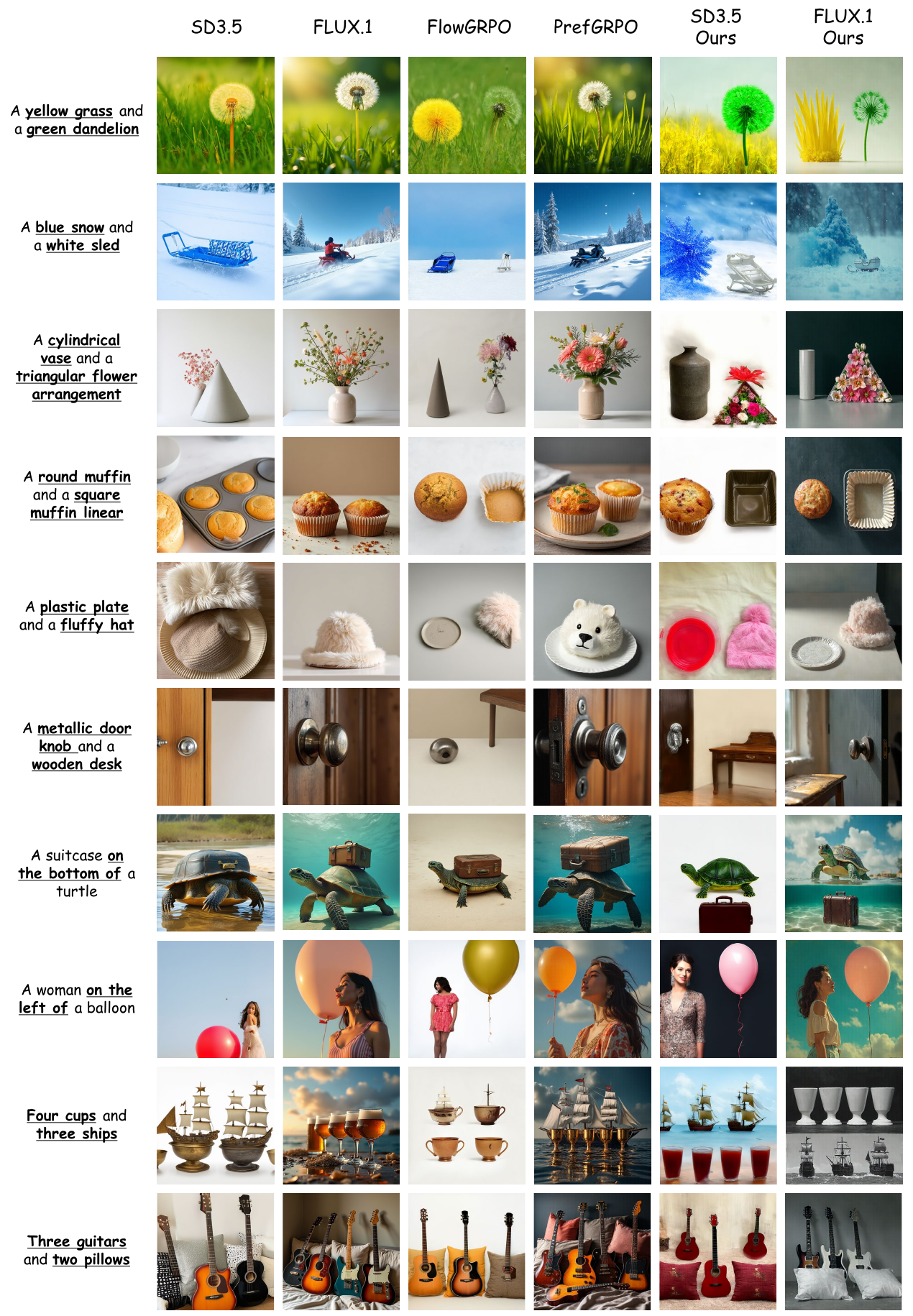}
    \caption{\textbf{Qualitative Comparison on T2I-CompBench.} Compared to baselines, \ours successfully handles multi-concept compositional prompts (e.g., attribute binding, spatial relations, numeracy) while effectively mitigating concept mis-alignment and missing objects.}
    \label{fig:t2i_qualitative}
\end{figure}

Figure~\ref{fig:t2i_qualitative} provides additional qualitative comparisons on T2I-CompBench. As illustrated, while baseline models frequently suffer from concept misalignment and object omission when faced with multi-concept prompts, \ours successfully handles diverse multi-concept prompts. By effectively generating precise attribute bindings, spatial relations, and numeracy, our method demonstrates superior alignment and robustness.

\section{Training Cost}
\label{sec:cost}
\begin{table}[htbp]
\centering
\scriptsize 
\caption{\textbf{Training Efficiency and Cost Comparison.} Comparison of training costs and configurations grouped by base model architectures. Our method achieves highly efficient fine-tuning using significantly fewer compute resources.}
\label{tab:training_cost_grouped}
\begin{tabular}{lccccc}
\toprule
\textbf{Base Model} & \textbf{Trans. Size} & \textbf{TE Size} & \textbf{Method} & \textbf{Tuning Method} & \textbf{GPUs} \\
\midrule
\multirow{2}{*}{SD3.5-M~\cite{esser2024scaling}} & \multirow{2}{*}{2B} & \multirow{2}{*}{5B} & FlowGRPO~\cite{liu2025flow} & LoRA & 24 \\
 & & & \textbf{\ours} & LoRA & \textbf{4} \\
\midrule
\multirow{2}{*}{FLUX.1-dev~\cite{flux2024}} & \multirow{2}{*}{12B} & \multirow{2}{*}{5B} & PrefGRPO~\cite{Pref-GRPO&UniGenBench} & FFT & 64 \\
 & & & \textbf{\ours} & LoRA & \textbf{4} \\
\midrule
Qwen-Image~\cite{wu2025qwen} & 20B & 8B & - & - & - \\
\bottomrule
\end{tabular}
\vspace{-1em}
\end{table}
\noindent\textbf{Training Cost.}
As summarized in Table \ref{tab:training_cost_grouped}, we evaluate the training cost of our proposed method compared to existing RL-based fine-tuning approaches. For the SD3.5-M base model, while FlowGRPO requires 24 GPUs for LoRA fine-tuning, \ours{} achieves highly efficient optimization using only 4 GPUs. The computational advantage of our approach becomes even more pronounced when scaling to the significantly larger FLUX.1-dev model (12B transformer). While the baseline PrefGRPO relies on computationally prohibitive Full Fine-Tuning (FFT) across 64 GPUs, \ours{} successfully adapts the model using LoRA on the same 4-GPU setup. 
\begin{figure}[htbp]
    \centering
    \includegraphics[width=\linewidth]{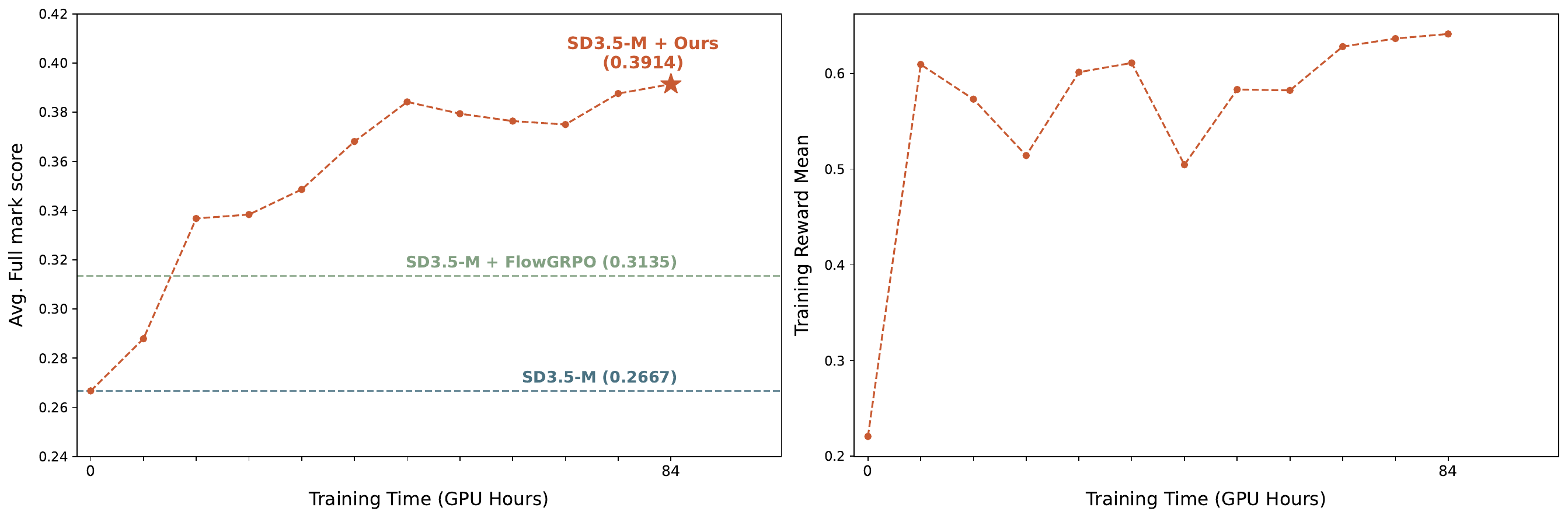}
    \caption{\textbf{Training Efficiency.} Avg. Full Mark score (left) and Training Reward Mean (right) over training time (GPU Hours). Our proposed method (SD3.5-M + Ours) demonstrates remarkable training efficiency, rapidly surpassing both the base SD3.5-M and the FlowGRPO baselines. Notably, the steady increase in training reward directly correlates with the consistent improvement in the ConceptMix evaluation.}
    \label{fig:training_time}
\end{figure}
This dramatic reduction in computational requirements highlights the exceptional efficiency and accessibility of our approach, making RL alignment for massive diffusion models highly feasible even in resource-constrained environments.

\noindent\textbf{Training Curve.} To evaluate the training efficiency of our proposed method, we track the Avg. Full Mark score across training time, as illustrated in Figure~\ref{fig:training_time}. The base model, SD3.5-M, achieves an average score of 0.2667. While FlowGRPO improves this score to 0.3135, it requires a significant training budget of over 2000 GPU hours~\cite{liu2025flow}. In contrast, our method (SD3.5-M + Ours) reaches a peak Avg. Full Mark score of 0.3914 using a total of only 84 GPU hours. Leveraging the efficient sampling of MixGRPO, our approach achieves rapid optimization, swiftly surpassing both baselines at the very early stages of training. Furthermore, we observe that as the training reward mean steadily increases (right), the ConceptMix score also consistently rises, demonstrating a strong correlation between our reward optimization and actual compositional performance. This highlights that our approach is exceptionally cost-effective and computationally efficient compared to previous online RL fine-tuning paradigms.

\section{SD3.5-M Results on ConceptMix}
\label{sec:addational_res}
\begin{figure}[htbp]
    \centering
    \includegraphics[width=\linewidth]{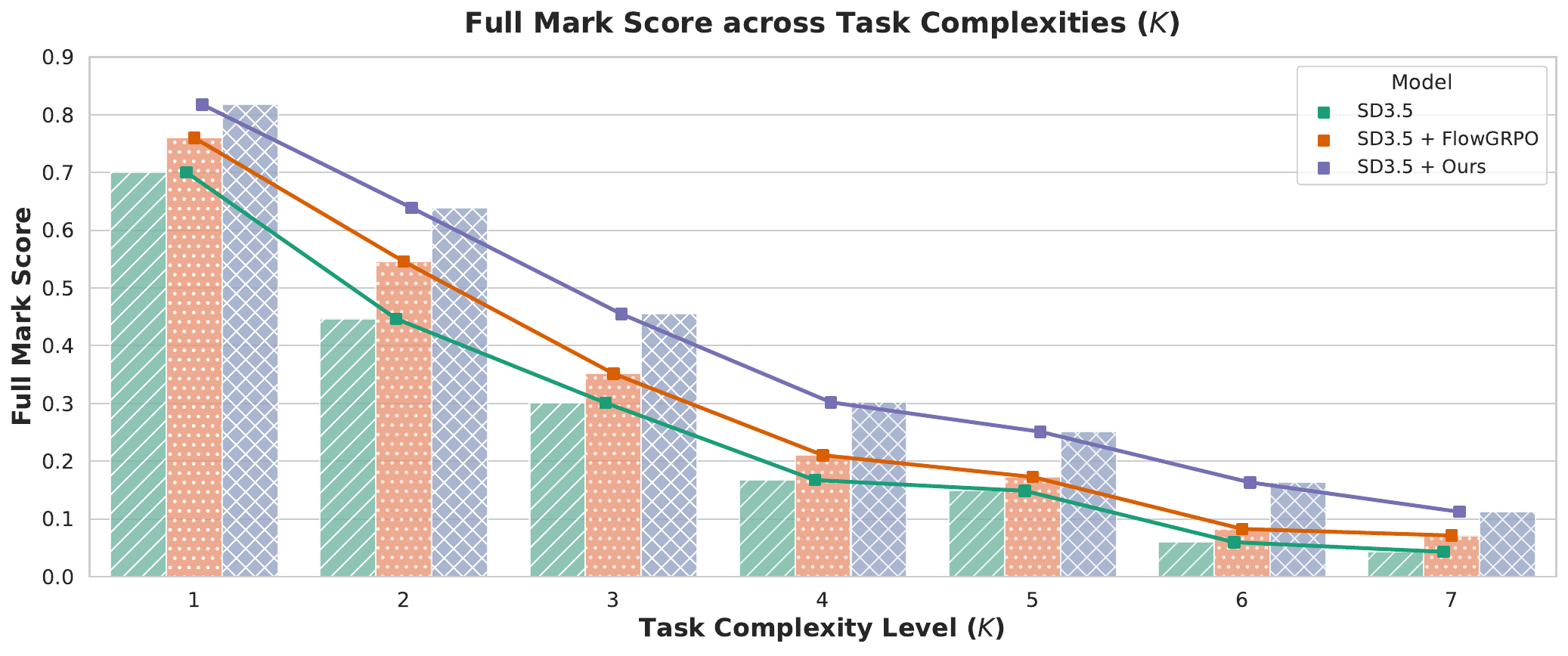}
    \caption{\textbf{Robustness against Task Complexity.} Comparison of Full Mark score across varying Task Complexity Levels ($K$) for SD3.5-M, SD3.5-M + FlowGRPO, and our proposed method (\ie, CMO). Our model consistently outperforms the baselines across all complexity levels.}
    \label{fig:fullmark_score}
\end{figure}

As illustrated in Figure \ref{fig:fullmark_score}, we evaluate the Full Mark score of the models across varying task complexity levels ($K$). Generally, as the number of concepts ($K$) increases from 1 to 7, the performance of all models exhibits a steady declining trend. However, our proposed method, denoted as SD3.5-M + Ours, consistently outperforms both the base SD3.5-M and the SD3.5-M + FlowGRPO baselines across all complexity levels. Notably, even at higher complexity levels ($K \ge 4$), our model maintains a relatively robust generation capability compared to the baselines, suffering from a less severe performance degradation. This demonstrates that our approach effectively preserves accurate object generation and prompt alignment capabilities, even under highly complex, multi-concept conditions.

\section{Baselines}
\label{sec:baselines_appendix}
To comprehensively validate the effectiveness of \ours, we compare it against a diverse set of representative compositional T2I baselines, categorized by their architecture and alignment strategies:
\begin{itemize}
\item \textbf{Autoregressive Models:} We include high-capacity autoregressive generators such as Janus~\cite{wu2025janus}, Show-o~\cite{xie2024show}, and HermesFlow~\cite{yang2025hermesflow}, which have recently shown strong performance in vision-language tasks.
\item \textbf{Diffusion-based Models:} We evaluate standard diffusion backbones including various iterations of Stable Diffusion (v1.4~\cite{rombach2022high}, v2~\cite{podell2023sdxl}, XL~\cite{podell2023sdxl}, XL Turbo~\cite{sauer2024adversarial}, 3.5~\cite{esser2024scaling}), PixArt-$\alpha$~\cite{chen2023pixart}, Playground V2.5~\cite{li2024playground}, and FLUX.1-dev~\cite{flux2024}.
\item \textbf{Alignment and RL-based Methods:} Our primary comparison focuses on recent state-of-the-art alignment methods, including IterComp~\cite{zhang2024itercomp}, FlowGRPO~\cite{liu2025flow}, and PrefGRPO~\cite{Pref-GRPO&UniGenBench}. We also include large-scale models such as Qwen-Image~\cite{wu2025qwen} and NanoBanana~\cite{comanici2025gemini}.
\end{itemize}
To demonstrate backbone-agnostic scalability, we report results for both SD3.5-M and FLUX.1-dev backbones. For general benchmarks, we primarily utilize the high-capacity FLUX.1-dev as the reference model to evaluate against large-scale baselines.

\end{document}